\documentclass[preprint,12pt]{elsarticle}
\usepackage{amssymb}
\usepackage{placeins}
\usepackage{rotating}
\usepackage[table]{xcolor}
\usepackage{lscape}
\usepackage{longtable}
\newcolumntype{L}{>{\centering\arraybackslash}m{6cm}}
\newcolumntype{M}{>{\centering\arraybackslash}m{2.5cm}l}

\biboptions{sort&compress}

\journal{International Journal of Medical Informatics}

\begin{document}

\begin{frontmatter}

\title{Generalizable Machine Learning for Stress Monitoring from Wearable Devices: A Systematic Literature Review}

\author[inst1]{Gideon Vos}

\affiliation[inst1]{organization={College of Science and Engineering, James Cook University},
            addressline={James Cook Dr}, 
            city={Townsville},
            postcode={4811}, 
            state={QLD},
            country={Australia}}

\author[inst1]{Kelly Trinh}
\author[inst2]{Zoltan Sarnyai}
\author[inst1]{Mostafa Rahimi Azghadi}

\affiliation[inst2]{organization={College of Public Health, Medical, and Vet Sciences, James Cook University},
            addressline={James Cook Dr}, 
            city={Townsville},
            postcode={4811}, 
            state={QLD},
            country={Australia}}

\begin{abstract}
\paragraph{Introduction}
Wearable sensors have shown promise as a non-intrusive method for collecting biomarkers that may correlate with levels of elevated stress. Stressors cause a variety of biological responses, and these physiological reactions can be measured using biomarkers including Heart Rate Variability (HRV), Electrodermal Activity (EDA) and Heart Rate (HR) that represent the stress response from the Hypothalamic-Pituitary-Adrenal (HPA) axis, the Autonomic Nervous System (ANS), and the immune system. While Cortisol response magnitude remains the gold standard indicator for stress assessment \cite{Boucher2019}, recent advances in wearable technologies have resulted in the  availability of a number of consumer devices capable of recording HRV, EDA and HR sensor biomarkers, amongst other signals. At the same time, researchers have been applying machine learning techniques to the recorded biomarkers in order to build models that may be able to predict elevated levels of stress. 

\paragraph{Objective}
The aim of this review is to provide an overview of machine learning techniques utilized in prior research with a specific focus on model generalization when using these public datasets as training data. We also shed light on the challenges and opportunities that machine learning-enabled stress monitoring and detection face.  

\paragraph{Methods}
This study reviewed published works contributing and/or using public datasets designed for detecting stress and their associated machine learning methods. The electronic databases of Google Scholar, Crossref, DOAJ and PubMed were searched for relevant articles and a total of 33 articles were identified and included in the final analysis. The reviewed works were synthesized into three categories of publicly available stress datasets, machine learning techniques applied using those, and future research directions. For the machine learning studies reviewed, we provide an analysis of their approach to results validation and model generalization. The quality assessment of the included studies was conducted in accordance with the IJMEDI checklist \cite{Cabitza2021}.

\paragraph{Results}
A number of public datasets were identified that are labeled for stress detection. These datasets were most commonly produced from sensor biomarker data recorded using the Empatica E4 device, a well-studied, medical-grade wrist-worn wearable that provides sensor biomarkers most notable to correlate with elevated levels of stress. Most of the reviewed datasets contain less than twenty-four hours of data, and the varied experimental conditions and labeling methodologies potentially limit their ability to generalize for unseen data. In addition, we discuss that previous works show shortcomings in areas such as their labeling protocols, lack of statistical power, validity of stress biomarkers, and model generalization ability.

\paragraph{Conclusion}
Health tracking and monitoring using wearable devices is growing in popularity, while the generalization of existing machine learning models still require further study, and research in this area will continue to provide improvements as newer and more substantial datasets become available.

\end{abstract}

\begin{graphicalabstract}
\begin{center}
  \makebox[\textwidth]{\includegraphics[width=\paperwidth]{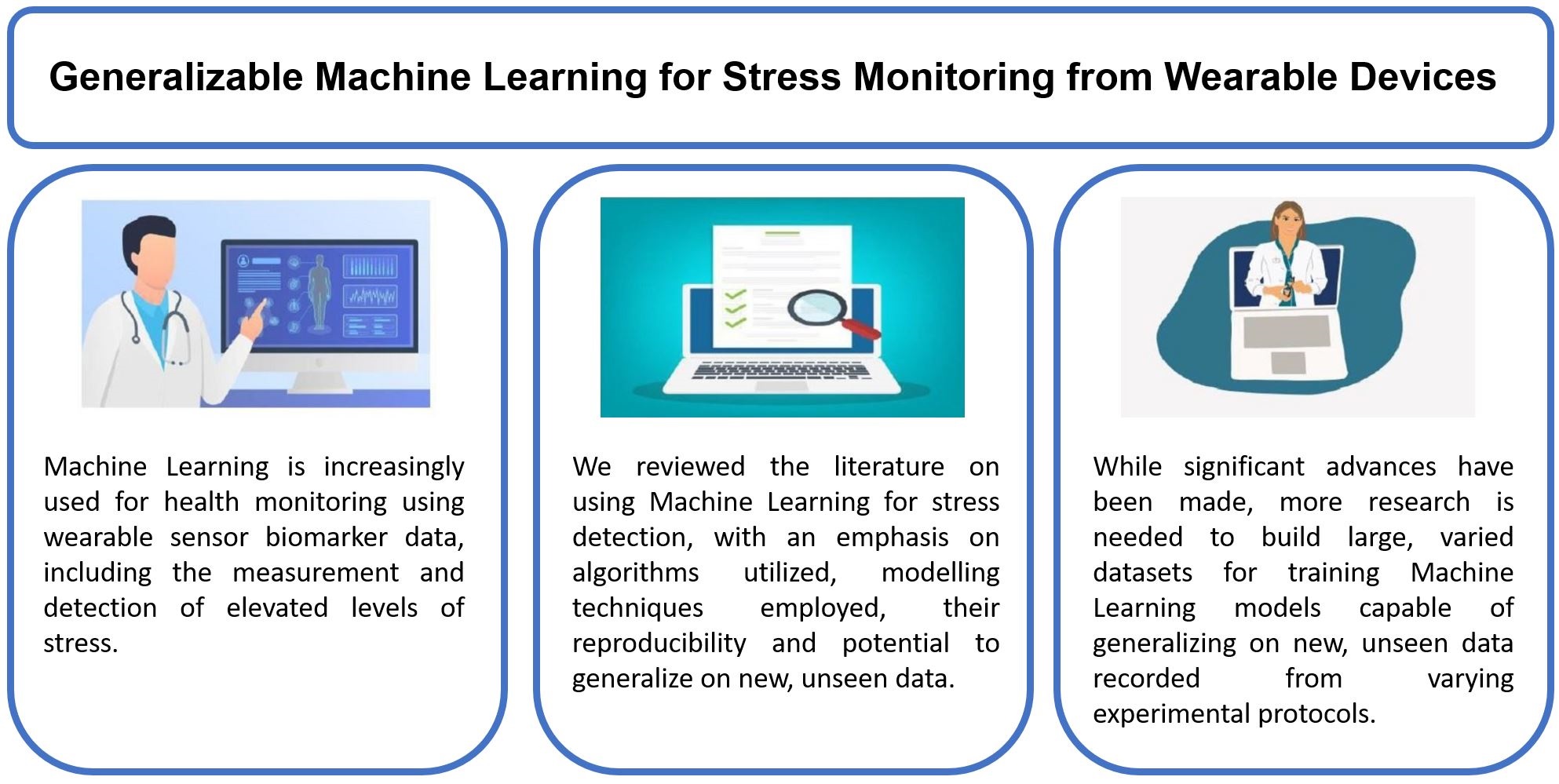}}
\end{center}
\end{graphicalabstract}

\begin{highlights}
\item This paper provides a review of the current state of stress detection
and measurement from wearable devices using machine learning, performed in accordance with the IJMEDI checklist. 
\item Our review of machine learning models is provided by analyzing and synthesizing the literature based on model generalization and reproducibility on unseen data.
\item We show that most stress-related machine learning studies are performed on small, singular datasets with a lack of generalization, and larger studies that utilize substantially more varied datasets
are needed.
\end{highlights}

\begin{keyword}
Stress \sep Wearable Sensor \sep Machine learning \sep Generalization
\PACS 07.05.Mh \sep 87.85.fk
\MSC 68T01 \sep 92C99
\end{keyword}

\end{frontmatter}


\section{Introduction}

\noindent Stress can be defined as the body’s psychological and physiological response to physical, emotional or mental strain. Such change in the environment elicits the activation of a cascade of biological responses (stress response) in the brain and in the body \cite{McEwen1998}. The stress response serves an important evolutionary role of helping the adaptation of the organism to the dynamically changing external and internal environment. This is achieved through mobilization of energy and its appropriate redistribution to organs that most immediately serve the adaptational response. At present, a universally recognized standard for stress evaluation remains outstanding \cite{Kim2018}, further compounded with the need for a comprehensive framework for investigating how organisms function in and adapt to constantly changing environments \cite{Thayer2012}. In the context of this paper and its reviewed studies, stress is considered as a binary condition for prediction. The data in a number of these studies \cite{Schmidt2018, Kraaij2015} was labeled with binary stressed or non-stressed time periods, and models trained on these datasets resulted in classifiers that would predict an observation as either stressed or not-stressed, while the other datasets  \cite{Rossi2020} utilized a daily stress inventory score \cite{Brantley1987} and one dataset \cite{Haouij2018} was labeled through observer scoring (0 to 1, low to high). A single study by Siirtola \emph{et al.} \cite{Siirtola2020} investigated and compared models trained as classifiers to models trained as a regression, where a threshold was established by analyzing the obtained continuous prediction values study subject-wise to obtain a balanced accuracy rate is as high as possible. In the studies reviewed, no single thresholding method could be determined that can generalize well across models.\\

\noindent Interestingly, a growing number of studies are examining the effects of training machine learning models on biomarker data collected in a study setting compared to daily life scenarios \cite{Can2020}, with further studies examining the effect of context when both training and evaluating predictive power \cite{Gjoreski2017g}. While the majority of studies in this review approached the training of machine learning models for stress detection as a single time-series dataset, more studies are evaluating the potential of person-specific models \cite{Nkurikiyeyezu2019} compared to generic non-specific models, with person-specific models showing great promise as powerful predictors of stress.\\

\noindent Wearable devices for personal health monitoring and tracking have gained significant popularity and technical sophistication since the release of the first Fitbit \cite{Fitbit2022} in 2009 and Empatica Embrace model in 2016 \cite{Empatica2022}. Recently, more advanced devices including Empatica's E4 \cite{Empatica2022} have been developed that are capable of measuring a wide variety of physiological signals. Peake \emph{et al.} \cite{Peake2018} performed a critical review of available wearable devices for providing bio-feedback, monitoring stress, and sleep with a critical review of their technical characteristics, reliability and validation. Continuous measurement of the physiological signals recorded using wearables enable researchers to extract useful information from these devices to potentially detect and monitor a variety of health-related events such as seizures \cite{Regalia2019,Tang2021,Onorati2021}, dehydration \cite{Kulkarni2020}, cognitive load \cite{Gjoreski2020}, physical activity \cite{Gholamiangonabadi2020}, emotions \cite{Zhang2020} and specifically related to this review, stress \cite{Gjoreski2020,Kaczor2020g, Gjoreski2017g,  Siirtola2020, Smets2016, Alshamrani2021g, Iqbal2021g, Can2019, Indikawati2020g, Nkurikiyeyezu2019, Han2020, Greco2021g, Sevil2020g}.\\

\noindent A number of previous survey articles have studied the topics of stress detection using wearable devices \cite{Samson2020} and machine learning \cite{Gedam2021}. In particular, in \cite{Samson2020}, Samson and Koh have surveyed various stress biomarkers and their measurement tools including wearables for salivary and electrochemical detection. However, they have not discussed how machine learning can be used to help with stress detection and measurements. In \cite{Gedam2021}, Gedam and Paul have surveyed works that have performed stress detection using wearable sensors measuring Electrocardiogram (ECG), Electroencephalography (EEG), and Photoplethysmography (PPG) signals and surveyed machine learning techniques for that. However, in this paper, we systematically review studies that have mainly used biomarker data from medical-grade wearable devices available to the consumer, due to the growing popularity of personal health monitoring, different to those used in \cite{Gedam2021}.\\

\noindent In addition, the previous reviews have not addressed a number of important points such as the statistical power \cite{Riley2020} of the training data used or their labeling protocols, and how it may affect machine learning model performance. Neither have they considered machine learning model generalization, where models built on any of the available public stress datasets are capable of accurately measuring stress when applied on a new dataset, or applied on datasets recorded under different conditions including experimental set-up, session duration, and labeling methodology.\\

\noindent Towards addressing these questions, we first explore the current state of stress detection and measurement using medical-grade wearable devices that are available to the consumer. We further explore the available public datasets built using sensor data recorded from these devices, and investigate the approaches utilized, and detection accuracy scores attained for machine learning models trained on these datasets. Finally, we discuss the generalization ability and limitations of these machine learning models, in order to understand the current state of using wearable devices for accurately measuring stress response and future directions.\\

\section{Methods}\label{sec:methods}
\subsection{Research questions}
\noindent The main aim of this work is to provide an overview of the current state of stress detection using machine learning techniques by using the IJMEDI checklist to assess the quality of the included literature, and specifically the generalization ability of models trained on public stress biomarker datasets and the potential reproducibility of their findings and results. Thus, our research questions can be formulated as follows:

\begin{itemize}
\item \textbf{RQ1:} Which machine learning algorithms and techniques are being utilized and trained on publicly available stress biomarker datasets?
\item \textbf{RQ2:} What accuracy metrics are reported and how are these findings being validated? Are the findings reproducible and does the methods utilized show promise towards model generalization?
\end{itemize}

\noindent Answering these questions will aid in getting a better understanding of the most current and accurate machine learning models available for predicting stress using wearable devices, and assist towards building a model capable of generalization on new, unseen data.\\

\subsection{Search strategy}
\begin{figure}[h!]
\centering
\includegraphics[width=0.8\textwidth]{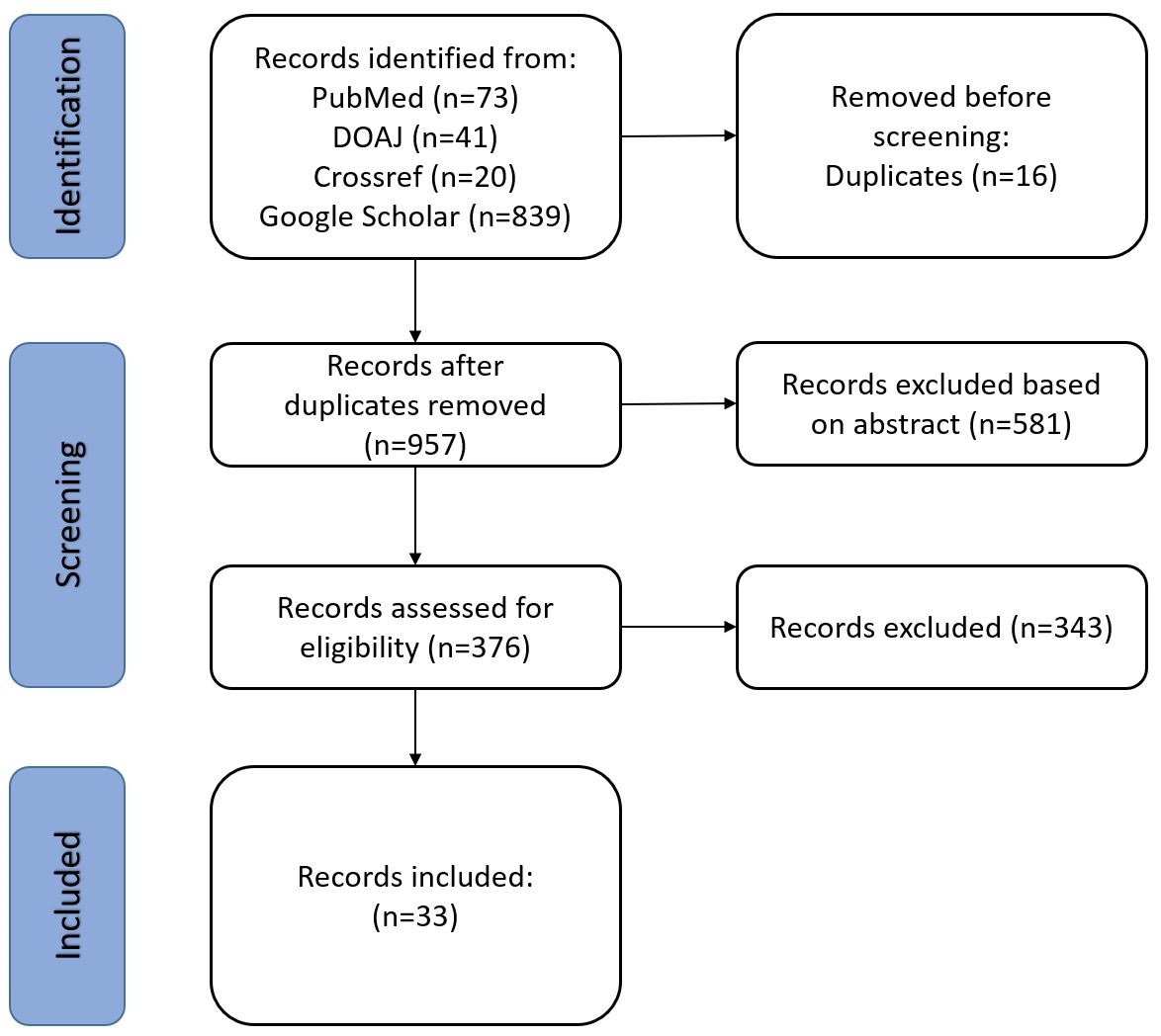}
\caption{\label{fig:figure1}Article screening process and the intermediate counts.}%
\end{figure}

\noindent We reviewed key published works (Figure \ref{fig:figure1}) between 2012 and 2022 on publicly available datasets related to stress, and more specifically, recorded using wearable devices; and measuring and predicting stress response using machine learning. The electronic databases of Google Scholar, Crossref, DOAJ and PubMed were searched for relevant articles using the keywords \emph{stress}, \emph{machine learning} and \emph{wearable} in title or abstract, and a total of 973 papers were identified. Duplicates were identified, and 16 were found and removed, leaving the number of considered papers for the subsequent phases at 957. Abstracts were scanned and irrelevant papers were excluded, including papers where the full text was not available. A small number of papers, in which the focus was stress in animals or psychiatry, were excluded. Studies using devices that are generally considered as health-trackers, or lifestyle monitors were also excluded, as were studies performed solely using devices that would not generally be considered a wearable device, such as EEG or chest-worn monitors. We further limited this review to machine learning models trained on, and devices that are capable of, recording multiple biomarkers that are known to be robust indicators of elevated levels of stress, i.e. HRV, EDA, HR, Inter-beat Interval (IBI) \cite{Samson2020}. Finally, papers where key machine learning techniques including feature-engineering and model validation techniques were not detailed, were also removed. As a result, a total of 33 papers were chosen for the systematic review process, grouped by the high-level topics of: \emph{Datasets}, \emph{Machine Learning for Stress Detection} and \emph{Future Research and Open Problems}. Table \ref{tab:papersreviewed} details the papers included in this review.

\subsection{Assessment of the quality of the studies}
\noindent Two reviewers (Vos and Azghadi) used the IJMEDI checklist \cite{Cabitza2021} to evaluate the quality of the included studies independently. The IJMEDI checklist is a quality assessment tool for medical artificial intelligence studies proposed by the IJMEDI, which aims to distinguish high-quality machine learning studies from simple medical data-mining studies. Six dimensions are included as 30 questions in the checklist: problem and data understanding, data preparation, modeling, validation, and deployment. Each question can be answered as OK (adequately addressed), mR (sufficient but improvable), and MR (inadequately addressed). In high-priority items, OK, mR and MR were assigned the scores of 0, 1, and 2, respectively, whereas in low priority items, the scores were halved. The maximum possible score was 50 points, with study quality was divided into low (0–19.5), medium (20–34.5), and high (35–50).

\begin{landscape}
\fontsize{10}{11}\selectfont
\begin{longtable}{ p{2.8cm} p{1.5cm} p{10cm} p{1cm} }
\caption{\label{tab:papersreviewed}Studies included in this review.}\\
\hline\hline
\textbf{Topic}                  &      \textbf{Reference}               & \textbf{Paper}                                                                                                                                                                                                                                                                                             & \multicolumn{1}{l}{\textbf{Date}}                                \\
\hline
\hline\endhead  

\hline\endfoot  

\rowcolor[rgb]{0.753,0.753,0.753} Data Sets        & \cite{Kraaij2015}            & The swell knowledge work dataset for stress and user modeling research                                                                                       & 2015                                                                                     \\
Data Sets        & \cite{Schmidt2018}           & Introducing WESAD, a multimodal dataset for wearable stress and affect detection                                                                             & 2018                                                                                     \\
\rowcolor[rgb]{0.753,0.753,0.753} Data Sets        & \cite{Haouij2018}            & AffectiveROAD system and database to assess driver's attention                                                                                               & 2018                                                                                     \\
Data Sets        & \cite{Gjoreski2020}          & Datasets for cognitive load inference using wearable sensors and psychological traits                                                                        & 2020                                                                                   \\
\rowcolor[rgb]{0.753,0.753,0.753} Data Sets        & \cite{Rossi2020}             & Multilevel monitoring of activity and sleep in healthy people                                                                                                & 2020                                                                                       \\
Data Sets        & \cite{Park2020}              & K-emocon, a multimodal sensor dataset for continuous emotion recognition in naturalistic conversations                                                       & 2020                                                                                 \\
\rowcolor[rgb]{0.753,0.753,0.753} Data Sets        & \cite{Svoren2020}            & Toadstool: A dataset for training emotional intelligent machines playing super mario bros                                                                    & 2020                                                                                    \\
Data Sets & \cite{Iqbal2022a}             & Comparison of machine learning techniques for psycho-physiological stress detection                                                                           & 2022                                                                                  \\

\rowcolor[rgb]{0.753,0.753,0.753} Machine Learning & \cite{Smets2018}             & Comparison of machine learning techniques for psycho-physiological stress detection                                                                           & 2016                                                                                   \\
Machine Learning & \cite{Gjoreski2017g}         & Monitoring stress with a wrist device using context                                                                                                          & 2017                                                                \\
\rowcolor[rgb]{0.753,0.753,0.753} Machine Learning & \cite{Can2019}               & Continuous stress detection using wearable sensors in real life: Algorithmic programming contest case study                                                  & 2019                                                                                          \\
Machine Learning & \cite{Kaczor2020g}           & Objective Measurement of Physician Stress in the Emergency Department Using a Wearable Sensor                                                                & 2020                                                                                     \\
\rowcolor[rgb]{0.753,0.753,0.753} Machine Learning & \cite{Indikawati2020g}       & Stress Detection from Multimodal Wearable Sensor Data                                                                                                        & 2020                                                                                       \\
Machine Learning & \cite{Sevil2020g}            & Detection and Characterization of Physical Activity and Psychological Stress from Wristband Data                                                             & 2020                                                                                        \\
\rowcolor[rgb]{0.753,0.753,0.753} Machine Learning & \cite{Mishra2020g}           & Evaluating the Reproducibility of Physiological Stress Detection Models                                                                                      & 2020                                                                                    \\
Machine Learning & \cite{Gholamiangonabadi2020} & Deep neural networks for human activity recognition with wearable sensors: Leave-one- subject-out cross-validation for model selection                       & 2020                                                                                   \\
\rowcolor[rgb]{0.753,0.753,0.753} Machine Learning & \cite{Jin2020}               & Predicting stress in teens from wearable device data using machine learning methods                                                                          & 2020                                                                                       \\
Machine Learning & \cite{Siirtola2020}          & Comparison of regression and classification models for user-independent and personal stress detection                                                      & 2020                                                                                          \\
\rowcolor[rgb]{0.753,0.753,0.753} Machine Learning & \cite{Gjoreski2020}          & Datasets for cognitive load inference using wearable sensors and psychological traits                                                                        & 2020                                                                                  \\
Machine Learning & \cite{Delmastro2020}         & Cognitive Training and Stress Detection in MCI Frail Older People Through Wearable Sensors and Machine Learning                                              & 2020                                                                                    \\
\rowcolor[rgb]{0.753,0.753,0.753} Machine Learning & \cite{Can2020}               & How Laboratory Experiments Can Be Exploited for Monitoring Stress in the Wild: A Bridge Between Laboratory and Daily Life                                    & 2020                                                                                          \\
Machine Learning & \cite{Elgendi2020}           & Machine Learning Ranks ECG as an Optimal Wearable Biosignal for Assessing Driving Stress                                                                     & 2020                                                                                     \\
\rowcolor[rgb]{0.753,0.753,0.753} Machine Learning & \cite{Garg2021}              & Stress Detection by Machine Learning and Wearable Sensors                                                                                                    & 2021                                                                                  \\
Machine Learning & \cite{Dalmeida2021}          & HRV Features as Viable Physiological Markers for Stress Detection Using Wearable Devices                                                                     & 2021                                                                                          \\
\rowcolor[rgb]{0.753,0.753,0.753} Machine Learning & \cite{Iqbal2021g}            & A Sensitivity Analysis of Biophysiological Responses of Stress for Wearable Sensors in Connected Health                                                      & 2022                                                                                     \\
Machine Learning & \cite{Alshamrani2021g}       & An Advanced Stress Detection Approach based on Processing Data from Wearable Wrist Devices                                                                   & 2021                               \\
\rowcolor[rgb]{0.753,0.753,0.753} Machine Learning & \cite{Gedam2021}             & A review on mental stress detection using wearable sensors and machine learning techniques                                                                   & 2021                                                                                      \\
Machine Learning & \cite{Greco2021g}            & Acute stress state classification based on electrodermal activity modeling                                                                                   & 2021                                                          \\
\rowcolor[rgb]{0.753,0.753,0.753} Machine Learning & \cite{Liapis2021}            & Advancing Stress Detection Methodology with Deep Learning Techniques Targeting UX Evaluation in AAL Scenarios & 2021                                                                                       \\
Machine Learning & \cite{Ehrhart2022}           & A Conditional GAN for Generating Time Series Data for Stress Detection in Wearable Physiological Sensor Data                                                 & 2022                                                                                         \\
\rowcolor[rgb]{0.753,0.753,0.753} Machine Learning & \cite{Iqbal2022}             & Exploring Unsupervised Machine Learning Classification Methods for Physiological Stress Detection                                                            & 2022                                                               \\
Future Research  & \cite{Swan2013}              & The quantified self: Fundamental disruption in big data science and biological discovery                                                                     & 2013                                                                                       \\
\rowcolor[rgb]{0.753,0.753,0.753} Future Research  & \cite{Westerink2020}         & Deriving a cortisol-related stress indicator from wearable skin conductance measurements: Quantitative model  experimental validation                      & 2020                                \\
\hline\hline

\end{longtable}
\end{landscape}

\section{Results}\label{sec:results}

\subsection{Wearable Devices for Stress Measurement}

\noindent Advances in hardware such as component miniaturization have enabled more technological features to be embedded into ever shrinking devices at lower cost. However, adoption is clearly a challenge that demands the collaborative attention of healthcare providers, hardware and software engineers, data scientists, policy-makers, cognitive neuroscientists, device engineers and materials scientists, among other specializations \cite{Azghadi2020}. From the initial Fitbit device launched in 2009, through to the Empatica E4 and the latest Oura Ring 3, significant improvements have been realized in both base features, as well as capabilities specifically related to the monitoring of, and promise to assist in improving, the user's overall health.\\

\noindent There are a wide variety of wearable devices in the market  \cite{Peake2018} used for health monitoring, including both medical-grade devices (Empatica Embrace Plus, Empatica E4, NOWATCH, Oura Ring) and consumer-oriented devices (Apple iWatch, Fitbit, Garmin, Samsung Gear). Consumer-oriented devices generally provide web-based platforms and smartphone applications for reporting various health statistics and levels of stress, with no ability to extract raw biomarker sensor recordings for scientific study, in contrast to medical-grade devices, such as the Empatica range, that provides full biomarker data download and additional support for researchers to properly utilize the raw signals directly for study.\\

\noindent However, in this review, our focus was limited to devices that are capable of stand-alone monitoring, without the need for an additional harness or pairing with a secondary device (worn on the wrist, finger or arm), as this would limit the usefulness for study outside of a stricter laboratory setting. Table \ref{tab:devices} provides a non-exhaustive list of well-known wearable devices potentially capable of tracking and monitoring stress.\\

\noindent Siirtola \cite{Siirtola2019} performed a study on smart watches reporting stress using a single biomarker (HR) and concluded that to be sufficient for detecting stress. Farrow \emph{et al.} \cite{Farrow2013} concluded that EDA is a robust, reliable, non-subjective psycho-physiological biomarker of psychological stress within subjects, but not always between. Greco \emph{et al.} \cite{Greco2021g} concluded that using only the EDA biomarker is sufficient for accurately predicting stress. The validity of sensor biomarkers is an open research question, discussed in detail in Section \ref{sensorvalidity}. Devices reporting stress based on only a single biomarker (typically HR or HRV) were therefor excluded.\\

\begin{table}
\centering
\caption{\label{tab:devices}Wearable devices for health tracking and monitoring.}
\resizebox{\textwidth}{!}{
\begin{tabular}{lccll}
\hline\hline
\textbf{Device} & \textbf{Release Year} & \textbf{Type} & \textbf{Sensors} & \textbf{Battery Life}  \\
                   \hline
\rowcolor[rgb]{0.753,0.753,0.753}NOWATCH                & 2023                                      & Wrist         & HR, TEMP, SpO2, EDA           & 2 Weeks                \\
Empatica Embrace Plus  & 2022                                      & Wrist         & EDA, ACC, TEMP, PR, PRV, ACC  & 1 Week                 \\
\rowcolor[rgb]{0.753,0.753,0.753}Fitbit Sense 2         & 2022                                      & Wrist         & HR, TEMP, SpO2, EDA           & 6 Days                 \\
Oura Ring 3            & 2021                                      & Finger        & HR, TEMP, SpO2, EDA           & 1 Week                 \\
\rowcolor[rgb]{0.753,0.753,0.753}Samsung Galaxy Watch 3 & 2020                                      & Wrist         & BVP, HR, ACC                  & 48 Hours               \\
Apple Watch 7          & 2019                                      & Wrist         & HR, ACC, SpO2                 & 18 Hours               \\
\rowcolor[rgb]{0.753,0.753,0.753}Fossil Gen 5           & 2019                                      & Wrist         & BVP, HR, ACC                  & 24 Hours               \\
Garmin Fenix 6X Pro    & 2019                                      & Wrist         & BVP, HR, ACC, SpO2            & 21 Days                \\
\rowcolor[rgb]{0.753,0.753,0.753}Polar OH1              & 2019                                      & Arm           & BVP, ACC                      & 12 Hours               \\
Fitbit Charge 3        & 2018                                      & Wrist         & HR, ACC                       & 1 Week                 \\
\rowcolor[rgb]{0.753,0.753,0.753}Garmin VivoActive 3    & 2018                                      & Wrist         & HR, ACC                       & 1 Week                 \\
Study Watch            & 2017                                      & Wrist         & HR, TEMP, EDA                 & 1 Week                 \\
\rowcolor[rgb]{0.753,0.753,0.753}Moodmetric             & 2017                                      & Finger        & EDA                           & 1 Week                 \\
Empatica E4            & 2015                                      & Wrist         & HR, TEMP, SpO2, EDA, ACC, IBI & 48 Hours               \\
\rowcolor[rgb]{0.753,0.753,0.753}Sony SmartBand 2       & 2015                                      & Wrist         & BVP, HR, ACC                  & 10 Hours               \\
Samsung Gear Live      & 2014                                      & Wrist         & BVP, HR, ACC                  & 24 Hours               \\
\rowcolor[rgb]{0.753,0.753,0.753}Philips DTI-2          & 2014                                      & Wrist         & EDA, ACC, TEMP                & 30 Hours          \\
\hline 
\multicolumn{5}{l}{ACC - accelerometer, BVP - Blood volume pulse, EDA - Electrodermal activity} \\
\multicolumn{5}{l}{HR - Heart rate, IBI - Inter-beat Interval, PR - Pulse rate} \\
\multicolumn{5}{l}{PRV - Pulse rate variability, SpO2 - Oxygen saturation, TEMP - Temperature}
\end{tabular}
}
\end{table}

\noindent The studies included in this review predominantly utilized datasets that are publicly available and therefor available to other researchers, and of these, the predominant wearable device utilized was the Empatica E4. Patient privacy when utilizing public health data for wearable research remain a concern, and Differential Privacy (DP) has emerged as a proficient technique to publish privacy sensitive data, including data from wearable devices. Saifuzzaman \emph{et al.} \cite{Saifuzzaman2022} conducted a Systematic Literature Review to identify, select and critically appraise research in DP to understand the different techniques available in wearable data publishing, and proposed a number of solutions for protecting patient privacy. Of the public datasets reviewed and included in this study, all patient identifiable information were excluded from the datasets.\\

\noindent Additionally, the measurement of stress in people with mental disorders or intellectual disabilities is of growing interest. Simons \emph{et al.} \cite{Simons2021} presented a specific protocol for studying patterns of physiological stress in patients with challenging behaviour. However, in this review we found the vast majority of current studies were performed using data captured from predominantly healthy subjects, screened for a number of health conditions prior to inclusion. Table \ref{tab:health} lists the studies included in this review where health screening was explicitly noted in the study, or where a dataset was utilized that was built using biomarker data from subjects screened for inclusion based on reported health status.

\begin{table}[h!]
\centering
\caption{\label{tab:health}Reported health screening criteria of reviewed studies.}
\resizebox{\textwidth}{!}{
\begin{tabular}{llccccccc}
\hline\hline
\textbf{Paper}             & \textbf{Dataset}                                  & \textbf{Healthy} & \textbf{Pregnancy} & \textbf{Smoking} & \textbf{Caffeine} & \textbf{Alcohol} & \textbf{Mental Disorders} & \textbf{Cardiovascular Disease}  \\
\hline
\rowcolor[rgb]{0.753,0.753,0.753}\cite{Iqbal2021g}   & WESAD             & $\bullet$               & $\bullet$                  & $\bullet$                &                   &                  & $\bullet$                         & $\bullet$                                \\
\cite{Nkurikiyeyezu2019} & WESAD   & $\bullet$                & $\bullet$                  & $\bullet$                &                   &                  & $\bullet$                         & $\bullet$                                \\
\rowcolor[rgb]{0.753,0.753,0.753}\cite{Delmastro2020}  & Custom         & $\bullet$                &                    &                  &                   &                  & $\bullet$                         &                                  \\
\cite{Ehrhart2022}  & Custom             & $\bullet$                &                    &                  &                   &                  &                           &                                  \\
\rowcolor[rgb]{0.753,0.753,0.753}\cite{Schmidt2018}   & WESAD            & $\bullet$                & $\bullet$                  & $\bullet$                &                   &                  & $\bullet$                         & $\bullet$                                \\
\cite{Liapis2021}   & WESAD              & $\bullet$                & $\bullet$                  & $\bullet$                &                   &                  & $\bullet$                         & $\bullet$                                \\
\rowcolor[rgb]{0.753,0.753,0.753}\cite{Alshamrani2021g}  & WESAD      & $\bullet$                & $\bullet$                  & $\bullet$                &                   &                  & $\bullet$                         & $\bullet$                                \\
\cite{Greco2021g}  & Custom                & $\bullet$                &                    & $\bullet$                & $\bullet$                 & $\bullet$                &                           &                                  \\
\rowcolor[rgb]{0.753,0.753,0.753}\cite{Indikawati2020g}  & WESAD      & $\bullet$                & $\bullet$                  & $\bullet$                &                   &                  & $\bullet$                         & $\bullet$                                \\
\cite{Smets2016}   & Custom                & $\bullet$                &                    &                  &                   &                  & $\bullet$                         &                                  \\
\rowcolor[rgb]{0.753,0.753,0.753}\cite{Iqbal2022}   & SWELL                & $\bullet$                &                    &                  &                   &                  &                           & $\bullet$                                \\
\cite{Garg2021} & WESAD                    & $\bullet$                & $\bullet$                  & $\bullet$                &                   &                  & $\bullet$                         & $\bullet$                   \\
\hline\hline
\end{tabular}
}
\end{table}
\FloatBarrier

\subsection{Wearable Device Datasets for Stress Measurement}
\noindent A number of datasets are publicly available containing sensor data recorded using a variety of devices matching our inclusion criteria, as detailed in Table \ref{tab:datareviewed}. The reviewed datasets contain the biomarkers predominantly utilized for stress detection, specifically EDA and HR signals. Apart from the Toadstool dataset, all recorded sessions exceed 60 minutes. The AffectiveROAD and Toadstool datasets contain biomarkers for a relatively small sample size of 10 subjects each, and small sample sizes of 25 subjects or less is a common feature of all public datasets reviewed. The largest public dataset included for review, Stress-Predict \cite{Iqbal2022a}, contains biomarker data recorded using an Empatica E4 device for 35 test subjects.\\

\noindent Labeling of the included datasets were performed using one of two methods: (i) periodic, where specific time frames during the experiment were either labeled as stressed or non-stressed, while the test subject was placed under that perceived condition (a stressful test or action, or non-stressed, restful period), or (ii) scored as experiencing stress or no stress during a particular period, either by completing a self-scoring evaluation, or by an observer who perceived a level of stress by observing the emotional reaction of the subject during that period.\\

\begin{sidewaystable}[!t]
\setlength\tabcolsep{1pt}
\centering
\caption{\label{tab:datareviewed}Summary of reviewed public wearable device stress-related datasets.}
\resizebox{\textwidth}{!}{
\begin{tabular}{lcccccLLL}
\hline\hline
\textbf{Dataset}                                & \multicolumn{1}{l}{\textbf{Year |}} & \multicolumn{1}{l}{\textbf{Subjects |}} & \multicolumn{1}{l}{\textbf{Female |}} & \multicolumn{1}{l}{\textbf{Male |}} & \textbf{Duration |} & \textbf{Biomarkers |}                          & \textbf{Devices |}                                                             & \textbf{Labeling/Scoring}                                       \\
\hline
\rowcolor[rgb]{0.753,0.753,0.753} SWELL                                           & 2014                              & 25                                    & 8                                   & 17                                & 138 min           & EDA, HRV, ECG                                & Facial expression, body postures, Mobi                                   & Periodic: Neutral, Time Pressure, Interruptions                           \\
Neurological Status                             & 2017                              & 20                                    &                &              & 31 min            & ACC,EDA, TEMP, HR, SPO2                      & Empatica E4                                                                  & Periodic: Relax, Physical Stress, Emotional Stress, Relax, Emotional Stress, Relax \\

\rowcolor[rgb]{0.753,0.753,0.753} WESAD         & 2018                              & 15                                    & 3                                   & 12                                & 120 min           & ACC, EDA, BVP, IBI, HR, TEMP, ECG, EMG, RESP & RespiBAN, Empatica E4                                                        & Periodic: Preparation, Baseline, Amusement, Stress, Meditation, Recovery  \\
AffectiveROAD & 2018                              & 10                                    & 5                                   & 5                                 & 118 min           & EDA, HR, TEMP                                & Empatica and Zephyr BioHarness 3.0 chest belt                                & Scored by observer                  \\
\rowcolor[rgb]{0.753,0.753,0.753} Toadstool                                       & 2020                              & 10                                    & 5                                   & 5                                 & 50 min            & ACC, EDA, BVP, IBI, HR, TEMP                 & Empatica E4                                                                  & Periodic: Game play under time pressure                                   \\
MMASH         & 2020                              & 22                                    &               &             & 24 hrs            & ACC, EDA, BVP, IBI, HR, TEMP , Cortisol      & Empatica E4                                                                  & Daily Stress Inventory value (DSI)                              \\
\rowcolor[rgb]{0.753,0.753,0.753} K-EmoCon                                        & 2020                              & 32                                    & 12                                  & 20                                & 120 min           & ACC, EDA, BVP, IBI, HR, TEMP, EEG            & Empatica E4, Polar H7 Bluetooth Heart Rate Sensor, NeuroSky MindWave Headset & Self-report and observer scoring                        \\
Stress-Predict         & 2022                              & 35                                    &               &             & 60 mins            & BVP, HR, RR      & Empatica E4                                                                  & Periodic                              \\

\hline\hline
\end{tabular}
}
\end{sidewaystable}
\FloatBarrier

\noindent The American Psychological Association defines three types of stress - Acute, Episodic Acute and Chronic, further divided into Absolute Stressors (stressors that everyone exposed to them would interpret as being stressful) and Relative Stressors (stressors that only some exposed to them would interpret as being stressful). Albrecht \cite{Albrecht1979} further defined four common types of stress, namely Time Stress, Anticipatory Stress (concerns about future events), Situational Stress (situations that you have no control over) and Encounter Stress (worry about interacting with a certain person or group of people).\\

\noindent Table \ref{tab:stresstypes} provides a summary of the types of stressors applied during each study reviewed in this paper, as defined by Albrecht \cite{Albrecht1979}, with a number of studies including all four types within their study setting and protocol. All studies involved cognitive or work-related tasks under pressure, and as noted in Table \ref{tab:health}, study subjects were screened for known health conditions in virtually all studies. Of the studies reviewed, three collected stress biomarker data during normal life conditions.\\

\noindent Jin \emph{et al.} \cite{Jin2020} provided Empatica E4 devices to study subjects after device use training, allowing subjects to utilize the device event marker to indicate periods during the day when they felt moderate to high levels of stress. Kaczor \emph{et al.} \cite{Kaczor2020g} performed a similar study in an healthcare emergency department, while Can \emph{et al.} \cite{Can2020} investigated the predictive performance of models trained under laboratory conditions when predicting on data collected in normal life conditions, and found that models trained on data recorded during laboratory sessions outperformed models trained on data collected during normal daily life conditions, when predicting for daily life conditions. This particular study \cite{Can2020} is of importance to researchers interested in building models from study data, for use on patient data collected during normal life conditions.\\

\begin{table}[h!]
\centering
\caption{\label{tab:stresstypes}Summary of reviewed studies and study stressors types applied.}
\resizebox{\textwidth}{!}{
\begin{tabular}{llcccc}
\hline\hline
\textbf{Paper}                                               & \textbf{Dataset} & \textbf{Time Stress} & \textbf{Anticipatory Stress} & \textbf{Situational Stress} & \textbf{Encounter Stress}  \\
\hline
\rowcolor[rgb]{0.753,0.753,0.753} Ehrhart 2022 \cite{Ehrhart2022}             & Custom           &                      & $\bullet$                            & $\bullet$                           &                            \\
Iqbal 2022 \cite{Iqbal2022}                 & SWELL            & $\bullet$                    & $\bullet$                            &                             &                            \\
\rowcolor[rgb]{0.753,0.753,0.753} Iqbal 2021 \cite{Iqbal2021g}                & WESAD            & $\bullet$                    & $\bullet$                            & $\bullet$                           & $\bullet$                          \\
Liapis 2021 \cite{Liapis2021}               & WESAD            & $\bullet$                    & $\bullet$                            & $\bullet$                           & $\bullet$                          \\
\rowcolor[rgb]{0.753,0.753,0.753} Alshamrani 2021 \cite{Alshamrani2021g}      & WESAD            & $\bullet$                    & $\bullet$                            & $\bullet$                           & $\bullet$                          \\
Greco 2021 \cite{Greco2021g}                & Custom           & $\bullet$                    & $\bullet$                            & $\bullet$                           &                            \\
\rowcolor[rgb]{0.753,0.753,0.753} Garg 2021 \cite{Garg2021}                   & WESAD            & $\bullet$                    & $\bullet$                            & $\bullet$                           & $\bullet$                          \\
Delmastro 2020 \cite{Delmastro2020}         & Custom           &                      & $\bullet$                            & $\bullet$                           &                            \\
\rowcolor[rgb]{0.753,0.753,0.753} Indikawati 2020 \cite{Indikawati2020g}      & WESAD            & $\bullet$                    & $\bullet$                            & $\bullet$                           & $\bullet$                          \\
Gjoreski 2020 \cite{Gjoreski2020}           & Custom           & $\bullet$                    &                              & $\bullet$                           &                            \\
\rowcolor[rgb]{0.753,0.753,0.753} Sevil 2020 \cite{Sevil2020g}                & Custom           & $\bullet$                    & $\bullet$                            & $\bullet$                           & $\bullet$                          \\
Kaczor 2020 \cite{Kaczor2020g}              & Custom           & $\bullet$                    & $\bullet$                            & $\bullet$                           & $\bullet$                          \\
\rowcolor[rgb]{0.753,0.753,0.753} Han 2020 \cite{Han2020}                     & Custom           & $\bullet$                    & $\bullet$                            & $\bullet$                           & $\bullet$                          \\
Jin 2020 \cite{Jin2020}                     & Custom           & $\bullet$                    & $\bullet$                            & $\bullet$                           & $\bullet$                          \\
\rowcolor[rgb]{0.753,0.753,0.753} Can 2020 \cite{Can2020}                     & Custom           & $\bullet$                    & $\bullet$                            & $\bullet$                           & $\bullet$                          \\
Nkurikiyeyezu 2019 \cite{Nkurikiyeyezu2019} & WESAD            & $\bullet$                    & $\bullet$                            & $\bullet$                           & $\bullet$                          \\
\rowcolor[rgb]{0.753,0.753,0.753} Can 2019 \cite{Can2019}                     & Custom           & $\bullet$                    & $\bullet$                            & $\bullet$                           &                            \\
Schmidt 2018 \cite{Schmidt2018}             & WESAD            & $\bullet$                    & $\bullet$                            & $\bullet$                           & $\bullet$                          \\
\rowcolor[rgb]{0.753,0.753,0.753} Smets 2016 \cite{Smets2016}                 & Custom           & $\bullet$                    &                              & $\bullet$                           &                      \\
\hline
\end{tabular}
}
\end{table}
\FloatBarrier

\subsection{Machine Learning Algorithms and Techniques for Stress Measurement using Wearable data}
\noindent Reviewing the literature, we found several machine learning techniques applied to detect elevated levels of stress using wearable devices. Table \ref{tab:modelsreviewed} lists the papers reviewed and the machine learning algorithms utilized. In the following subsections, we provide a discussion on the different steps of the machine learning pipelines utilized, and analyze how previous works have performed those steps, noting their strengths and limitations.

\begin{sidewaystable}[!t]
\centering
\caption{\label{tab:modelsreviewed}Summary of the machine learning models reviewed.}
\renewcommand{\arraystretch}{1.5}
\resizebox{\textwidth}{!}{
\begin{tabular}{ccccccccc}
\hline\hline
\textbf{Paper}                                                                                                                                                               & \textbf{Year} & \textbf{Model}                & \textbf{Dataset} & \textbf{Accuracy} & \textbf{Subjects} & \textbf{Features} & \textbf{Cross Validation}     & \textbf{Window}    \\
\hline
\cite{Smets2016}                                                                            & 2016                     & Bayesian networks             & Custom         & 84.60\%                          & 20                           & 22                           & LOSO                             & 30s, 29s overlap   \\
\rowcolor[rgb]{0.753,0.753,0.753}\cite{Gjoreski2017g}                                                                                                       & 2017                     & SVM                           & Custom         & 71.00\%                          & 5                            & 6                            & LOSO                             & 6min               \\
\cite{Schmidt2018}                                                                            & 2018                     & Random Forest, LDA, AdaBoost  & WESAD          & 93.00\%                          & 15                           & 82                           & LOSO                             & 0.25s, 5s, 60s     \\
\rowcolor[rgb]{0.753,0.753,0.753}\cite{Can2019}                                                     & 2019                     & Neural Network, Random Forest & Custom         & 97.92\%                          & 21                           & 17                           & 10-Fold                          & 2min, 20min        \\
\cite{Nkurikiyeyezu2019}                                                                & 2019                     & Random Forest, ExtraTrees     & WESAD, SWELL   & 93.90\%                          & 15, 25                       & 94                           & 10-Fold                          & 5min, 10min        \\
\rowcolor[rgb]{0.753,0.753,0.753}\cite{Siirtola2020}                                                      & 2020                     & Bagged tree based ensemble    & AffectiveROAD  & 82.30\%                          & 9                            & 119                          & LOSO                             & 60s, 0.5s overlap  \\
\cite{Gjoreski2020}                                                                      & 2020                     & XGBoost                       & Snake, CogLoad & 82.00\%                          & 23,23                        & 44                           & LOSO                             &                    \\
\rowcolor[rgb]{0.753,0.753,0.753}\cite{Sevil2020g}                                                             & 2020                     & LDA                           & Custom         & 98.30\%                          & 24                           & 2216                         & 10-Fold                          &                    \\
\cite{Kaczor2020g}                                                               & 2020                     & Naive Bays                    & Custom         & 64.50\%                          & 8                            & 30                           & 10-Fold                          & 20min              \\
\rowcolor[rgb]{0.753,0.753,0.753}\cite{Han2020}                                                                                      & 2020                     & K-nearest-neighbor            & Custom         & 94.55\%                          & 17                           & 25                           & 10-Fold                          & 60s                \\
\cite{Jin2020}                                                                             & 2020                     & Random Forest                 & Custom         & 89.40\%                          & 8                            & 756                          & 10-Fold                          &                    \\
\rowcolor[rgb]{0.753,0.753,0.753}\cite{Indikawati2020g}                                                                                                   & 2020                     & RF                            & WESAD          & 92.00\%                          & 15                           & 4                            & 60/40 Split                      & 0.25s              \\
\cite{Elgendi2020}                                                                    & 2020                     & Ensemble                      & SRAD           & 75.02\%                          & 17                           & 6                            &                                  &                    \\
\rowcolor[rgb]{0.753,0.753,0.753}\cite{Can2020}                                       & 2020                     & Random Forest                 & Custom         & 74.61\%                          & 14                           & 2                            & 10-Fold                          &                    \\
\cite{Delmastro2020}                                           & 2020                     & Adaboost                      & Custom         & 85.30\%                          & 9                            & 10                           & 10-Fold                          &                    \\
\rowcolor[rgb]{0.753,0.753,0.753}\cite{Iqbal2021g}                                                      & 2021                     & Logistic Regression           & WESAD          & 85.71\%                          & 14                           & 5                            & 14-Fold                          & 60s                \\
\cite{Greco2021g}                                                                                   & 2021                     & SVM, RF                       & Custom         & 94.62\%                          & 65                           & 14                           & LOSO                             &                    \\
\rowcolor[rgb]{0.753,0.753,0.753}\cite{Alshamrani2021g}                                                              & 2021                     & Neural Network                & WESAD          & 85.00\%                          & 15                           & 14                           & LOSO                             &                    \\
\cite{Garg2021}                                                                                                      & 2021                     & RF                            & WESAD          & 83.34\%                          & 15                           & 5                            & LOSO                             & 10s                \\
\rowcolor[rgb]{0.753,0.753,0.753}\cite{Dalmeida2021}                                                                   & 2021                     & Gradient Boosting             & SRAD           & 79.00\%                          & 17                           & 7                            &                                  &                    \\
\cite{Liapis2021} & 2021                     & Neural Network                & WESAD          & 97.40\%                          & 15                           & 36                           & UX Dataset                       &                    \\
\rowcolor[rgb]{0.753,0.753,0.753}\cite{Iqbal2022}                                                             & 2022                     & Random Forest                 & SRAD, SWELL    & 65.6\%, 75\% & 17, 25  & 2                            & 70/30 Split                      &                    \\
\cite{Ehrhart2022}                                                & 2022                     & Neural Network                & Custom         & 72.62\%                          & 35                           &          & Test on unseen & 16s                \\
\hline\hline
\end{tabular}
}
\end{sidewaystable}
\FloatBarrier

\subsubsection{Pre-processing}
\noindent Electronic sensors used in wearable devices for recording biomarkers differ widely, and subsequently operate and record on different sampling frequencies. For the Empatica E4, for instance, the EDA signal is sampled at 4 Hz, while the HR signal is sampled at 1 Hz. Recorded session data for both sensors will therefor differ in length, and researchers will have to pre-process the sensor data by down-sampling the EDA signal to 1 Hz to ensure a like for like timestamp match with the HR signal, and subsequently any stress metric label for the exact time period. In the studies reviewed, \cite{Nkurikiyeyezu2019, Jin2020, Gjoreski2020, Indikawati2020g, Ehrhart2022} specifically noted that down-sampling was applied on data used within their experiments.\\  

\noindent Due to varying experimental protocols and the ease of collection of non-stress samples, data is likely to be unbalanced with more non-stress samples versus stressed samples present in any given dataset. Therefore, another usual pre-processing step performed on wearable stress data is class balancing that can be done in different ways. For instance, Nkurikiyeyezu \emph{et al.} \cite{Nkurikiyeyezu2019} balanced the recorded sensor data by randomly discarding some samples from the majority (non-stressed) class, and further applied logarithmic, square root, and Yeo-Johnson transformations to ensure a Gaussian distribution, as required by their use of a linear regression model. Can \emph{et al.} \cite{Can2019} also performed class-balancing through random down-sampling of the majority class (non-stressed observations) to match the minority class (stressed observations).\\

\noindent As noted in Table \ref{tab:classbalancing}, neither up-sampling nor down-sampling techniques showed a substantial difference or improvement in predictive power, and this may be due to the lack of a proven strategy employed when selecting which observations to discard, potentially causing information loss \cite{Ali2015} of important biomarker data during the sampling process. Class balancing techniques all have varied benefits and risks, as noted in Table \ref{tab:classbalancing}, and to this extent, a number of methods have been proposed to improve class-balancing re-sampling techniques. Deng \emph{et al.} \cite{Deng2022} proposed a unified approach for multivariate time series classification when data is imbalanced, while Lee \emph{et al.} \cite{Lee2022} used a semi-supervised technique known as Active Learning to mitigate the effect of imbalanced class labels. Jiang \emph{et al.} \cite{Jiang2021} proposed a new oversampling method based on the classification contribution degree to deal with a number of shortcomings when using SMOTE (Synthetic Minority Oversampling Technique) \cite{Chawla2002}, such as oversampling from noisy points. A notable drawback of reliance on class balancing when dealing with highly imbalanced datasets such as the stress biomarker datasets included in this study, where the stressed period is generally the minority class, is reproducibility and generalizability on new, unseen data that may contain significant outliers and a different class distribution, depending on the study setting and protocol used during biomarker recording. Further research is required to identify robust techniques for dealing with these class imbalances in physiological biomarker datasets.

\begin{table}[h!]
\centering
\caption{\label{tab:classbalancing}Class balancing methods employed in reviewed studies.}
\renewcommand{\arraystretch}{1.5}
\resizebox{\textwidth}{!}{
\begin{tabular}{llcll}
\hline\hline
\textbf{Paper}        & \textbf{Class Balancing}       & \textbf{Accuracy} & \textbf{Benefit}  & \textbf{Risk}  \\
\hline
\rowcolor[rgb]{0.753,0.753,0.753}\cite{Sevil2020g}                & ADASYN \cite{He2008}      & 98.30\% & Reduces bias \cite{He2008}  & Generates minority outliers \cite{Brandt2021}                             \\
\cite{Can2019}                   & Upsampling of Minority Class
      & 97.92\% & Simple implementation \cite{Brandt2021}    & Promotes overfitting \cite{Leevy2018}                            \\
\rowcolor[rgb]{0.753,0.753,0.753}\cite{Nkurikiyeyezu2019} & Downsampling of Majority Class & 93.90\% & Simple implementation  \cite{Brandt2021} & Information loss \cite{Ali2015}                             \\
\cite{Jin2020}                & Random Sampling \cite{Ali2015}               & 89.40\%          & Reduces bias \cite{He2008}     & Explainability                      \\
\rowcolor[rgb]{0.753,0.753,0.753}\cite{Delmastro2020}         & SMOTE \cite{Chawla2002}     & 85.30\%    & Surpasses random sampling methods \cite{Elreedy2023}  & Introduces noise, Overgeneralization \cite{Jiang2021}
                          \\
\cite{Can2020}                     & Downsampling of Majority Class & 74.61\%       & Simple implementation \cite{Brandt2021}    & Information loss \cite{Ali2015}                     \\
\rowcolor[rgb]{0.753,0.753,0.753}\cite{Ehrhart2022}             & Data Augmentation              & 72.62\% & Can reduce bias \cite{Pastaltzidis2022} & Explainability  \\                           
\hline\hline
\end{tabular}
}
\end{table}
\FloatBarrier

\noindent Differences in data range, units and scale can be problematic for some machine learning algorithms and standardization is usually applied to scale the data to have a mean of 0 and a standard deviation of 1. Similarly, the goal of normalization is to change the values of numeric columns in the dataset to a common scale, without distorting differences in the ranges of values. In the context of stress detection, normalization and standardization were utilized by \cite{Nkurikiyeyezu2019, Gjoreski2020, Sevil2020g, Iqbal2021g, Dalmeida2021}, with \cite{Gjoreski2020} experimenting on both raw and standardized data, and finding that standardization offered improved predictive performance across all 10 machine learning algorithms tested. Another usual pre-processing step on biomedical signals such as stress-related biomarkers collected by wearable devices is filtering. This is done to reduce outliers and any potential noise. For instance, \cite{Schmidt2018} applied a 5 Hz low-pass filter on the raw EDA signal, \cite{Iqbal2021g} applied a high-pass filter on the raw EDA signal,  while \cite{Nkurikiyeyezu2019, Han2020, Alshamrani2021g} applied a 4Hz fourth-order Butterworth low-pass filter, followed by a moving average filter, to reduce outliers and remove noise from EDA sensor signals.\\ 

\subsubsection{Feature-Engineering}
\noindent A common technique for extracting 
useful features representing physiological time series data, is to summarize the changing features of the existing data using summary statistics. Guo \emph{et al.} \cite{Guo2020} performed a study to evaluate summary statistics as features for clinical prediction tasks, and found that commonly used combinations of summary statistics such as [min, max, mean] and [min, max, mean, standard deviation (std)] achieved good prediction results in most cases. However, they reported that skew and kurtosis, which reflect the shape of a distribution, performed poorly when used individually as features for prediction, but appeared frequently in the optimal combinations, indicating that they can play a role as supplemental information.\\

\noindent The techniques noted by Guo \emph{et al.} \cite{Guo2020} were frequently applied in the stress detection studies reviewed. Fourteen of the reviewed approaches \cite{Smets2016, Schmidt2018, Can2019, Nkurikiyeyezu2019, Han2020, Siirtola2020, Gjoreski2017g, Greco2021g, Sevil2020g, Iqbal2021g, Dalmeida2021, Can2020, Elgendi2020, Garg2021} utilized summary statistics of biomarkers using a sliding-window approach, ranging from 0.25 seconds in one experiment up to 20 minutes in others, with varying degrees of success. In \cite{Kreibig2010}, the author noted summary windows of 30 and 60 seconds are most often utilized, based on the hypothesis that this factor correlates with physiological response. Can \emph{et al.} \cite{Can2019} decomposed the phasic and tonic components of the EDA signal using a convex optimization approach, as the tonic component includes more long-term slow changes, whereas phasic components include faster (event-related) changes. Both \cite{Can2019} and \cite{Gjoreski2020} found that sliding windows ranging between 10 and 17.5 minutes produced better detection accuracy, with \cite{Can2019} further noting that different machine learning algorithms relied on different window sizes, an important factor to consider for future research.\\

\noindent Jin \emph{et al.} \cite{Jin2020} used the \emph{tsfresh} Python library to automatically generate 4536 features off their existing data and applied a Random Forest model as machine learning approach. To evaluate the performance of such a large number of features, the results were grouped around the key biomarkers (i.e. HR, EDA, TEMP), from which the features were engineered. Gjoreski \emph{et al.} \cite{Gjoreski2017g} used greedy step-wise selection to identify the top features considered most useful for their specific machine learning model, and further noted that when sensor-specific features are used, PPG-based features achieved higher predictive accuracy results, followed by the IBI and HR-based features. Iqbal \emph{et al.} \cite{Iqbal2021g} found features based on HR and respiratory rate to be the most important, while Dalmeida \emph{et al.} \cite{Dalmeida2021} focused their research specifically on HRV as a viable biomarker, and found HRV features to constitute good markers for stress detection. \\

\subsubsection{Algorithm Selection}

\noindent Of the 23 machine learning based stress detection studies reviewed, we noted the use of 16 different algorithms, including combinations of Logistic Regression (LR), Support Vector Machines (SVM), Decision Trees (DT), Random Forests (RF), Bayesian Networks (BN), Principal Component Analysis (PCA), Linear Discriminant Analysis (LDA), k-Nearest Neighbour (kNN), Multi-layer Perceptron (MLP), Multi-task learning (MTL), Adaboost, Naive Bayes (NB), Bagging, Gradient Boosting (GB) and Neural Networks (NN). Of these, SVM, RF and kNN were the most commonly used for stress detection, with tree-based models such as RF and GB generally delivering better predictive performance on supervised binary classification objectives.\\

\noindent A standard approach consists of selecting a small number of algorithms that may be suitable for the problem, train each and select the best performing model based on their final predictive accuracy. \cite{Nkurikiyeyezu2019} experimented on a single method (Random Forest) while \cite{Siirtola2019} used 13 different algorithms to test the  predictive accuracy of classification based models versus regression type models for predicting elevated levels of stress, of which Bagged Trees performed the best. Similarly, \cite{Can2019, Smets2016, Schmidt2018, Han2020, Gjoreski2017g, Kaczor2020g} utilized 5 to 7 different algorithms and compared the stress prediction accuracy of each, with the highest performing models listed in Table  \ref{tab:modelsreviewed}. Iqbal \emph{et al.} \cite{Iqbal2022} compared the performance of 7 supervised methods to 7 unsupervised methods and concluded that a careful selection of classification models is required when aiming to develop an accurate stress detection system, with unsupervised machine learning classifiers showing good performance in terms of classification accuracy.\\

\noindent Additionally, the predictions from a set of algorithms can be combined based on averaging, weighted-averaging or voting, to produce a final prediction (commonly known as model ensembling). This technique was specifically noted in experiments done by Gjoreski \emph{et al.} \cite{Gjoreski2017g}, Kaczor \emph{et al.} \cite{Kaczor2020g} and Elgendi \emph{et al.} \cite{Elgendi2020}.\\

\subsubsection{Hyperparameter Optimization}

\noindent Hyperparameters can be defined as the different parameter values used to control the learning process of a machine learning algorithm, and can have a significant effect on their performance. Hyperparameter optimization is the process of finding the right combination of algorithm parameter values to achieve maximum performance on the given dataset. Examples of hyperparameters are the number of estimators (trees) and maximum tree depth in the Random Forest algorithm. Due to the large number of parameters that require tuning in different algorithms, automated methods \cite{Elgeldawi2021} have been developed to scan the full parameter search space in a reasonable amount of time to determine the optimal combination.\\

\noindent Of the stress-related studies reviewed, we noted \cite{Smets2016} restricted the hyperparameter of the estimators count used in their Random Forest model to 20, while \cite{Jin2020} performed a grid search with estimators set at 500. In \cite{Nkurikiyeyezu2019}, the authors used 1,000 estimators while limiting the tree depth to 2, in order to limit the possibility of over-fitting. For the decision tree classification algorithms used by \cite{Schmidt2018}, information gain was used to measure the quality of splitting decision nodes, and the minimum number of samples required to split a node was set to 20. The number of base estimators was set to 100 for both of their utilized algorithms (Random Forest and AdaBoost). In another study, Han \emph{et al.} \cite{Han2020} did not specifically optimize hyperparameters, but built several kNN models with different parameter values for \emph{k} (1, 3, 5, 7, 9) and selected the best performing model from those. Sevil \emph{et al.} \cite{Sevil2020g} utilized Bayesian optimization techniques for feature-selection.\\

\noindent Unlike the aforementioned works, Gjoreski \emph{et al.} \cite{Gjoreski2020} tuned their model parameters by randomly sampling from distributions predefined by an expert. The models were then trained with the specific parameters and evaluated using cross-validation on the training data. The best performing model from the cross-validation was used to classify the test data. A systematic, well-defined hyperparameter optimization approach is crucial to improve the reproducibility of scientific studies and ensures that machine learning algorithms are tailored to the problem at hand. As noted by Can \emph{et al.} \cite{Can2019}, the performance of machine learning models may be dependent on an optimal selection of window size when generating summary statistics to engineer features, and this needs consideration when selecting hyperparameters for optimal predictive performance.\\

\subsubsection{Model Training and Validation}

\begin{figure}[h!]
\centering
\fbox{\includegraphics[width=\textwidth]{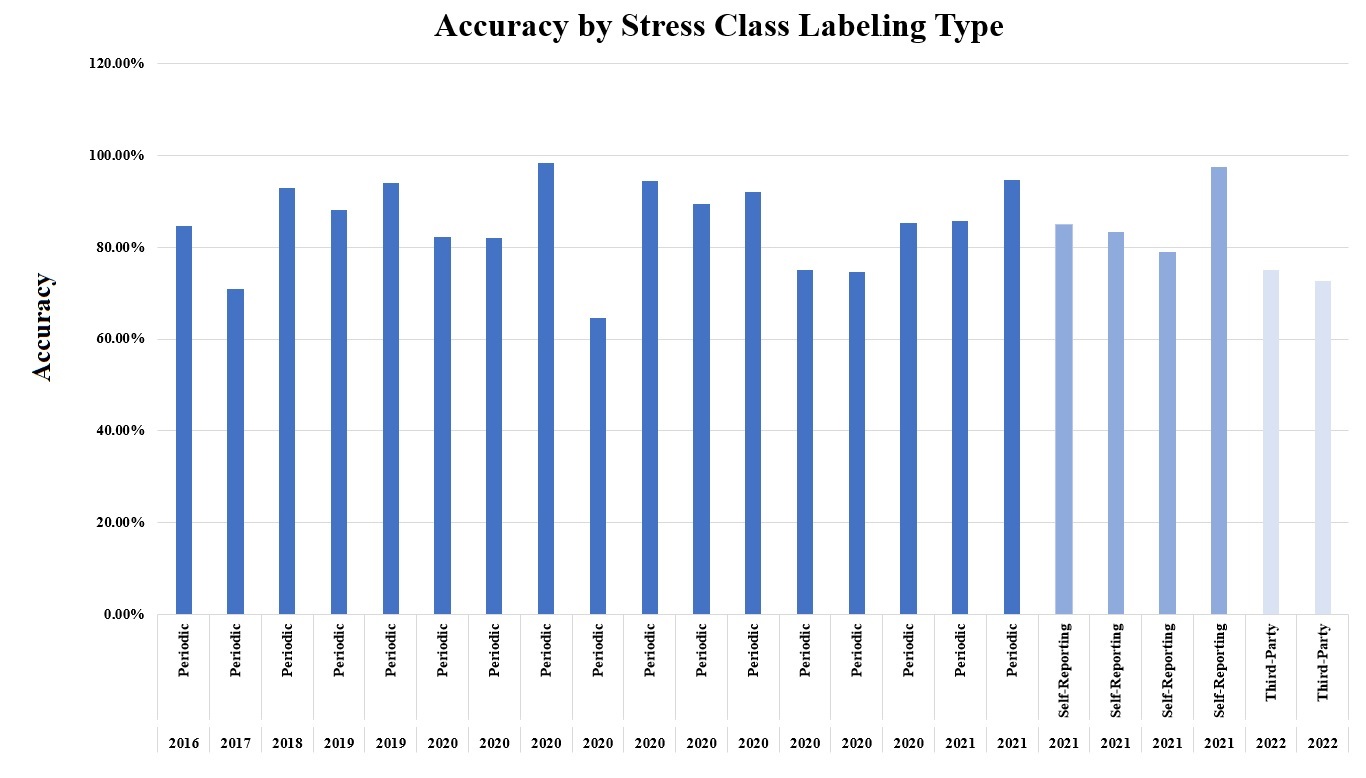}}
\caption{\label{fig:figure2}Accuracy based on labeling method included in study.}%
\end{figure}

\noindent An important requirement when developing supervised machine learning algorithms is to have valid labeled data. In the case of stress measurement, we found three main methods employed for labeling elevated levels of stress. These include (i) specific stress/no-stress periods marked during an experimental recording session \cite{Can2019, Kraaij2015, Svoren2020, Schmidt2018, Greco2021g, Kaczor2020g, Sevil2020g, Indikawati2020g, Alshamrani2021g, Iqbal2022, Liapis2021, Dalmeida2021, Garg2021, Elgendi2020, Ehrhart2022, Delmastro2020, Han2020,Iqbal2022a}; (ii) self-reporting via questionnaires \cite{Smets2016, Rossi2020, Park2020, Gjoreski2017g, Iqbal2021g, Gjoreski2020, Can2020}; and (iii) labeling by a third-party observer, who observes subjects' response to a situation and numerically scores/grades the level of stress observed \cite{Siirtola2020, Park2020, Haouij2018, Jin2020}. Figure \ref{fig:figure2} details the studies reviewed for each year, with reported accuracy rates by labeling method. Periodic labeling was the most commonly used labeling technique and provided consistently higher accuracy rates as reported by each study, compared to self-scoring and scoring by a third-party.\\

\noindent As highlighted in Table \ref{tab:modelsreviewed}, the best performing models from each experiment achieved at least 64.5\% test accuracy, with \cite{Schmidt2018, Nkurikiyeyezu2019, Han2020, Indikawati2020g, Greco2021g, Sevil2020g, Liapis2021} reporting binary classification test accuracy rates of over 90\%, using datasets labeled with specific, marked stress/no-stress periods. It should be noted that as stress is a physiological response, predictive accuracy in these experiments measure a predictive correlation between the included features (biomarkers) against a labeled metric at the same point in time (stressed versus non-stressed). Siirtola \emph{et al.} \cite{Siirtola2019} attempted to model how high this relationship is (using a regression algorithm instead of classification), while Umematsu \emph{et al.} \cite{Umematsu2020g} focused on the problem of forecasting future episodes of stress, rather than measuring levels of stress on previously recorded data.\\

\noindent Cross-validation is a re-sampling procedure used to evaluate machine learning models on a limited data sample. The purpose of cross–validation is to test the ability of a machine learning model to predict with high accuracy on new, unseen data. It is also used to flag problems like over-fitting or selection bias, and gives insights on how well the model will generalize to an independent dataset. Among the studies reviewed, \cite{Smets2016, Schmidt2018, Gjoreski2020, Siirtola2020, Gjoreski2017g, Alshamrani2021g, Greco2021g, Garg2021} utilized Leave One Subject Out (LOSO) cross-validation, while \cite{Can2019, Nkurikiyeyezu2019, Han2020, Jin2020, Kaczor2020g, Sevil2020g, Iqbal2021g, Delmastro2020, Can2020} utilized K-fold cross-validation with \emph{K=10}. In addition, \cite{Gjoreski2020} utilized both LOSO and K-fold cross-validation, with \emph{K=5}. All studies reviewed approached stress prediction as a binary classification problem apart from \cite{Siirtola2019}, where the problem type was defined as stress level measurement, rather than a binary stressed versus non-stressed problem. No definitive improvement in reported accuracy rates were noted when using LOSO cross-validation compared to K-fold cross-validation.\\

\subsubsection{Performance Analysis}

\noindent A wide variety of metrics are available for measuring machine learning model performance, depending on the problem being solved, for example classification or regression type problems. The experiments reviewed utilized and reported a number of different evaluation metrics including F-score \cite{Can2019, Gjoreski2020, Schmidt2018, Ehrhart2022, Iqbal2022, Garg2021}, classification accuracy \cite{Can2019, Han2020, Smets2016, Greco2021g, Kaczor2020g, Sevil2020g, Iqbal2021g, Indikawati2020g, Alshamrani2021g, Gjoreski2017g, Ehrhart2022, Delmastro2020, Iqbal2022, Can2020, Dalmeida2021, Elgendi2020, Garg2021}, Kappa \cite{Liapis2021}, Area Under the Curve (AUC) \cite{Jin2020, Dalmeida2021} and Mean Absolute Error (MAE) \cite{Nkurikiyeyezu2019}. For comparisons among the reviewed studies, here we only investigate their achieved classification accuracy, if reported.
Classification accuracy simply measures how often the classifier correctly predicts, i.e. what is the ratio of the number of correct predictions to the total number of predictions.\\

\noindent To determine classification accuracy, Kaczor \emph{et al.} \cite{Kaczor2020g} divided their dataset into 3 classes: a pre-stress event vs post-stress event, baseline vs pre-stress event, and baseline vs
post-stress event, reporting a classification accuracy rate of 90.4\%. Can \emph{et al.} \cite{Can2019} reported a binary classification accuracy rate of 97.92\%  using a custom (non-public) dataset, and Gjoreski \emph{et al.} \cite{Gjoreski2020} reported a binary classification accuracy rate of 68.2\% when applied on the \emph{CogLoad} dataset and 82.3\% when applied on the \emph{Snake} dataset. Han \emph{et al.} reported a binary accuracy rate of 94.55\%, while Nkurikiyeyezu \emph{et al.} \cite{Nkurikiyeyezu2019} and Schmidt \emph{et al.} \cite{Schmidt2018} reported binary classification accuracy rates of 93.9\% and 93\% respectively. Jin \emph{et al.} \cite{Jin2020} reported an AUC rate of 89.4\% rather than an accuracy rate. AUC, unlike classification accuracy, is sensitive to class imbalance when there is a minority class. This implies that classification accuracy rates can be high even if the predictions for a minority class is mostly wrong. This could lead to samples marked as non-stressed being classified mostly correctly and stressed samples (the minority class) predicted inaccurately, while still reporting an overall high accuracy rate.\\ 

\begin{figure}[ht!]
\centering
\fbox{\includegraphics[width=\textwidth]{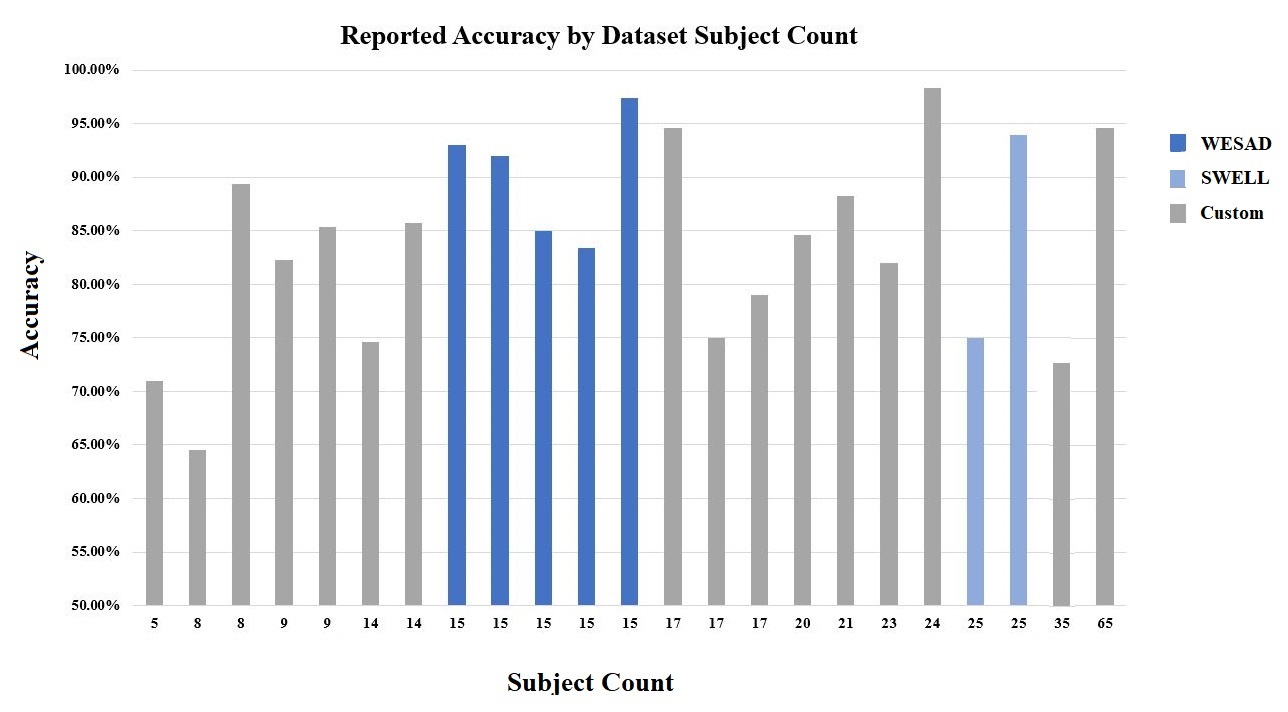}}
\caption{\label{fig:figure3}Accuracy based on number of subjects included in study.}%
\end{figure}

\noindent Figure \ref{fig:figure3} details the reported accuracy metrics achieved for the experiments reviewed in this paper, based on the size of the dataset used in terms of individual test subjects. The highest reported accuracy rate of 98.30\% was achieved by Sevil \emph{et al} \cite{Sevil2020g} when using Linear Discriminant Analysis (LDA) on a non-public dataset consisting of 24 test subjects, and validated using K-Fold cross-validation. Liapis \emph{et al.} \cite{Liapis2021} trained a Neural Network on the public WESAD dataset (15 subjects), and evaluated on a user-annotated dataset consisting of skin conductivity (SC) segments for 30 study participants, reporting an accuracy rate of 97.40\%.\\

\FloatBarrier

\subsubsection{Study Quality}

\begin{figure}[ht!]
\centering
\includegraphics[width=\textwidth]{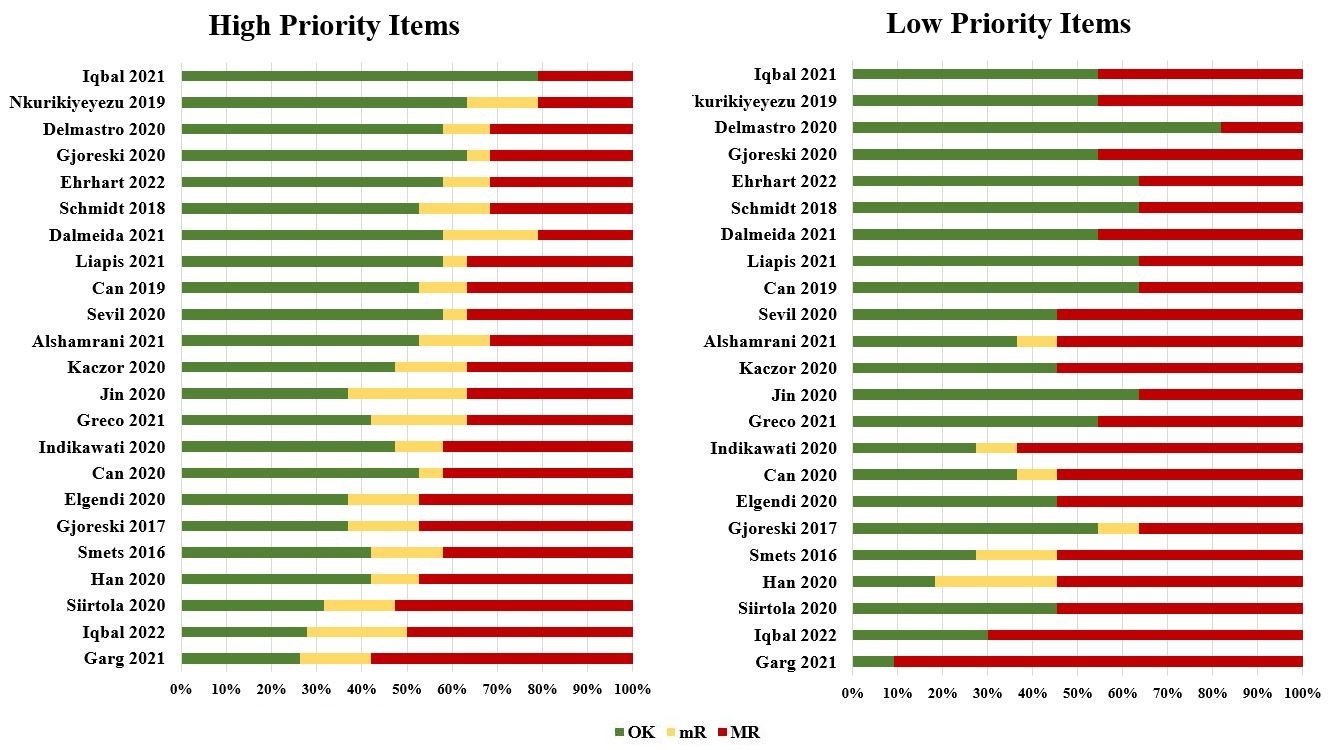}
\caption{\label{fig:figure4}Proportion of the different answers in the high- and low-priority items. OK = adequately addressed; mR = sufficient but improvable; MR = inadequately addressed.}%
\end{figure}

\noindent The \emph{Supplementary File} details the results of the IJMEDI quality assessment. Table \ref{tab:ijmedi} summarizes the scores of each dimension and the total score in each study. The average score of the included studies was 25.7 (range: 14–35). Most of the studies were of a medium quality, while one \cite{Iqbal2021} were of a high quality. The majority of the studies had an obvious bias in the quality of problem understanding, data understanding and modeling dimensions. Figure \ref{fig:figure4} shows the proportion of the different answers in the high and low priority items. 

\begin{sidewaystable}[!t]
\centering
\caption{\label{tab:ijmedi}Quality assessment scores of the 23 ML-based studies according to the IJMEDI checklist.}
\renewcommand{\arraystretch}{1.5}
\resizebox{\textwidth}{!}{
\begin{tabular}{lcccccccc}
\hline\hline
                   & \textbf{Problem Understanding (10)} & \textbf{Data Understanding (6)} & \textbf{Data Preparation (8)} & \textbf{Modeling (6)} & \textbf{Validation (12)} & \textbf{Deployment (8)} & \textbf{Total (5)} & \textbf{Generalization} \\
                   \hline
\rowcolor[rgb]{0.753,0.753,0.753} Iqbal 2021 \cite{Iqbal2021g}        & 7                                   & 6                               & 8                             & 6                     & 7                        & 2                       & 36         &         \\
Nkurikiyeyezu 2019 \cite{Nkurikiyeyezu2019} & 6                                   & 3                               & 8                             & 6                     & 8                        & 2                       & 33   &               \\
\rowcolor[rgb]{0.753,0.753,0.753} Delmastro 2020 \cite{Delmastro2020}    & 9                                   & 5                               & 2                             & 6                     & 7                        & 4                       & 33            &      \\
Gjoreski 2020 \cite{Gjoreski2020}     & 7                                   & 6                               & 2                             & 6                     & 8                        & 2                       & 31          &        \\
\rowcolor[rgb]{0.753,0.753,0.753} Ehrhart 2022 \cite{Ehrhart2022}      & 7                                   & 5                               & 2                             & 6                     & 9                        & 2                       & 31        & $\bullet$         \\
Schmidt 2018 \cite{Schmidt2018}      & 7                                   & 5                               & 1                             & 6                     & 8                        & 3                       & 30         &         \\
\rowcolor[rgb]{0.753,0.753,0.753} Dalmeida 2021 \cite{Dalmeida2021}     & 6                                   & 4                               & 4                             & 6                     & 8                        & 2                       & 30         &         \\
Liapis 2021 \cite{Liapis2021}       & 7                                   & 4                               & 2                             & 6                     & 9                        & 2                       & 30        & $\bullet$         \\
\rowcolor[rgb]{0.753,0.753,0.753} Can 2019 \cite{Can2019}          & 7                                   & 4                               & 2                             & 6                     & 7                        & 3                       & 29          &        \\
Sevil 2020 \cite{Sevil2020g}        & 5                                   & 4                               & 4                             & 6                     & 7                        & 2                       & 28          &        \\
\rowcolor[rgb]{0.753,0.753,0.753} Alshamrani 2021 \cite{Alshamrani2021g}    & 7                                   & 5                               & 4                             & 5                     & 5.5                      & 1                       & 27.5      &          \\
Kaczor 2020 \cite{Kaczor2020g}       & 9                                   & 3                               & 2                             & 5                     & 6                        & 1                       & 26         &         \\
\rowcolor[rgb]{0.753,0.753,0.753} Jin 2020 \cite{Jin2020}          & 6                                   & 3                               & 2                             & 6                     & 7                        & 2                       & 26       &           \\
Greco 2021 \cite{Greco2021g}        & 6                                   & 4                               & 0                             & 6                     & 8                        & 2                       & 26         & $\bullet$        \\
\rowcolor[rgb]{0.753,0.753,0.753} Indikawati 2020 \cite{Indikawati2020g}   & 7                                   & 5                               & 2                             & 6                     & 5.5                      & 0                       & 25.5       &         \\
Can 2020 \cite{Can2020}          & 7                                   & 4                               & 2                             & 6                     & 5.5                      & 1                       & 25.5        &        \\
\rowcolor[rgb]{0.753,0.753,0.753} Elgendi 2020 \cite{Elgendi2020}      & 7                                   & 2                               & 2                             & 6                     & 6                        & 1                       & 24           &       \\
Gjoreski 2017 \cite{Gjoreski2017g}     & 5                                   & 3                               & 2                             & 5                     & 5.5                      & 3                       & 23.5         &       \\
\rowcolor[rgb]{0.753,0.753,0.753} Smets 2016 \cite{Smets2016}        & 6.5                                 & 5                               & 0                             & 5                     & 5.5                      & 1                       & 23          &        \\
Han 2020 \cite{Han2020}          & 6.5                                 & 4                               & 0                             & 5                     & 5.5                      & 0.5                     & 21.5      &          \\
\rowcolor[rgb]{0.753,0.753,0.753} Siirtola 2020 \cite{Siirtola2020}     & 5                                   & 2                               & 1                             & 5                     & 6                        & 1                       & 20          &        \\
Iqbal 2022 \cite{Iqbal2022}  &6 & 2 & 0 & 6 & 5 & 0 & 19     &            \\
\rowcolor[rgb]{0.753,0.753,0.753} Garg 2021 \cite{Garg2021}         & 1                                   & 2                               & 0                             & 5                     & 6                        & 0                       & 14         &        \\
\hline\hline
\end{tabular}
}
\end{sidewaystable}
\FloatBarrier

\FloatBarrier

\section{Discussion}\label{sec:discussion}

\noindent In order to build a robust machine learning model capable of accurately detecting stress, we consider four important requirements. These include (i) Sensor biomarker data needs to be valid and sufficiently varied to capture a wide spectrum of potential physiological stress response; (ii) For supervised machine learning, this data needs to be accurately labeled where observations are marked as stressed or non-stressed or a stress score range is given, to allow the model to learn from the data; (iii) Where a specific hypothesis is being tested, a sufficient level of statistical power is required \cite{cohen1988spa, Brysbaert2019, Melvin2022, Riley2020, Iqbal2022a}, thereby ensuring results and findings can be considered statistically significant, and (iv) Model generalization occurs in order to apply the model on new, unseen data, with high accuracy. The discussion of this review is therefor focused on those four key requirements.\\

\noindent Having scored the machine learning studies included in this review using the IJMEDI checklist, we found only one study \cite{Iqbal2021} of high quality, with the remaining studies being of medium quality, and a single study being of low quality \cite{Garg2021}. Most studies scored well in problem, data understanding and modeling domains. Data preparation and deployment scores were notably low, as were scores for high priority items in the validation domain. Interestingly, there were no notable improvement in study quality over time.\\

\noindent Five studies \cite{Nkurikiyeyezu2019, Gjoreski2020, Delmastro2020, Iqbal2021, Ehrhart2022} scored over 30, being of medium to higher quality. Focusing on the modeling, validation and deployment domain scores of these eight studies, we note an improvement in quality over time for only the validation domain, indicating a lack of progress in the modeling domain and little focus on the deployment of models in real-life scenarios, including factors pertaining to sustainability, model bias and ethics.\\

\begin{figure}[ht!]
\centering
\fbox{\includegraphics[width=\textwidth]{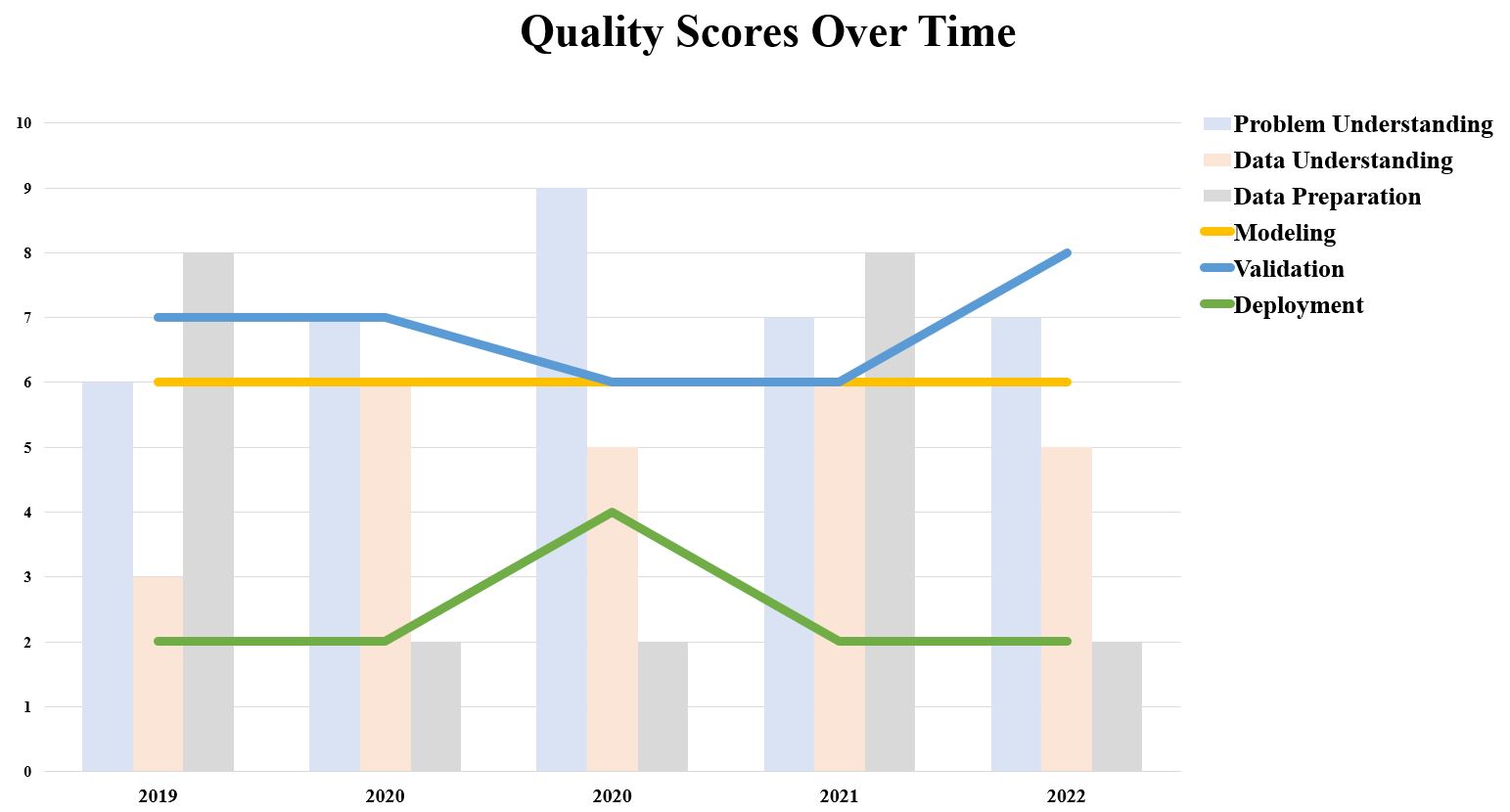}}
\caption{\label{fig:figure5}Trend of higher-quality studies over time.}%
\end{figure}
\FloatBarrier

\subsection{Validity of sensor biomarkers} \label{sensorvalidity}
\noindent The datasets included in this study contain a variety of sensor biomarker data potentially useful in detecting elevated levels of stress via HR, HRV and EDA signals, measured across a time interval. In addition, sweat sensing is at the forefront of wearable stress detection currently in development \cite{Samson2020} and devices sensing sweat may hold great promise to quantify several biomarkers, namely Cortisol, to monitor the levels of stress that an individual is experiencing.\\

\noindent However, \cite{Lier2019} noted that at present, there is a lack of consensus on a standardized protocol or framework with which to test the validity of physiological signals measured by these devices and their derived parameters. It is also argued that sudden, short-lived stressors, such as being startled by the ringing of the phone, or possible habituation effects as a result of exposure to repeated information cannot be validly detected. Lier \emph{et al.} \cite{Lier2019} reported that physiological changes during a workday can be tracked by the Empatica E4 wearable against more major, sustained stressors.\\

\noindent In \cite{Milstein2020}, the authors found the Empatica E4 to be suitable for psychotherapy research focused on Inter-Beat Interval (IBI) and specific HRV measures, but failed to produce reliable EDA data and produced missing IBI data, especially when a subject is being more dynamic. This is confirmed by Ryan \emph{et al.} \cite{Ryan2019} and Sevil \emph{et al.} \cite{Sevil2020g} that found the Empatica E4 can be severely compromised by motion artifact. This can result in a high percentage of missing data across all conditions except seated and supine baselines, and questions the E4's efficacy as an HRV measurement tool in most in-vivo conditions. This is further confirmed by Georgiou \emph{et al.} \cite{Georgiou2018} that found wearable devices can only be used as a surrogate for HRV at resting or mild exercise conditions, as their accuracy fades out with increasing exercise load, and Schuurmans \emph{et al.} \cite{Schuurmans2020} who noted the potential of the Empatica E4 as a practical and valid tool for research on HR and HRV under non-movement conditions. Seipaejaervi \emph{et al.} \cite{Seipaejaervi2022} found that an HRV-based stress index mirrors responses of cortisol, and an HRV-based stress index may be used to quantify physiological responses to psychosocial stress across various health and age groups. In contrast, Greco \emph{et al.} \cite{Greco2021g} found EDA to be a good marker of stress when features are engineered based on its phasic and tonic components.\\ 

\noindent In \cite{Siirtola2020}, the authors predominantly focused on comparing regression vs. classification models using the AffectiveROAD dataset \cite{Haouij2018}. This dataset contains sensor recordings for both left and right hands of the test subjects. To ensure consistent, comparable results, \cite{Siirtola2020} utilized only data recorded from the right hand of each test subject, leaving the important question of sensor placement unanswered, and needing further study to confirm whether sensor placement on the dominant versus non-dominant hand of a test subject could potentially affect biomarker accuracy, and more importantly for this review, correlation with increased levels of stress. Empatica note on their website \cite{Empatica2022} that newer studies have shown substantial differences in the EDA signal between the dominant and non-dominant hand.\\

\subsection{Labeling protocol}
\noindent In terms of labeling protocol and methodology,
\cite{Siirtola2020} questioned the accuracy of self-reporting of perceived levels of stress experienced, which was previously questioned by \cite{Schmidt2019}, who noted that study subjects are less likely to report on states less socially desired. Accurate labeling of stressed/non-stressed periods in the sensor data is crucial to building a reliable and robust machine learning model. To achieve this, in the datasets reviewed in Table \ref{tab:datareviewed}, two major labeling methods were used. The SWELL, Toadstool and WESAD datasets were recorded with specific intervals to denote stressed/non-stressed periods for labeling. In the AffectiveROAD, MMASH and K-EmoCon datasets, on the other hand, labeling was performed using self or observed stress indicator scoring. An interesting observation is that where these datasets were utilized in reviewed machine learning models, the models trained on periodically-labeled data achieved significantly higher levels of detection accuracy compared to the models trained using self or observed stress scoring. This is likely due to false negative reporting in the questionnaires, as noted by \cite{Schmidt2019}.\\

\noindent Stress is not a binary condition, and none of the studies reviewed noted specific methods for establishing thresholds within biomarkers to utilise as indicators of periods of high stress (low, moderate, high). Any potential thresholds established by the machine learning algorithms, (specifically tree-based methods) during training were not examined in detail to determine any potential time-varying dynamics between the biomarkers. Ghiasi \emph{et al.} \cite{Ghiasi2020} proposed combining HRV and EDA correlates as a single index, rather than treating each as separate indicators of ANS changes. They reported good results when validating this metric on two experimental protocols.\\ 

\subsection{Lack of Statistical Power}

\noindent Of the papers included in this review, six \cite{Smets2016, Nkurikiyeyezu2019, Elgendi2020, Delmastro2020, Iqbal2021g, Dalmeida2021} specifically included a hypothesis statement in their experiments. However, no power analysis were noted in any of the machine learning papers reviewed, regardless of hypothesis statement. It is common to design behavioral science experiments with a statistical power of 80\% or higher \cite{cohen1988spa}, which reduces the probability of encountering a Type II error by up to 20\% \cite{brownlee_2020}. Statistical power has three parts: effect size (a statistical measure), sample size (number of observations or participants) and significance (typically 0.05 \cite{cohen1988spa}). Power analysis assists researchers in determining the smallest sample size suitable to detect the effect of a given experiment at a desired level of significance, as collecting larger samples are likely costlier and much harder. The use of machine learning in behavioral science experiments does not automatically negate the need for sufficient statistical power \cite{Melvin2022, Riley2020}.\\

\noindent One of the recurrent questions psychology researchers ask is: \emph{"What is the minimum number of participants I must test?"} \cite{Brysbaert2019}. The high number of participants required for an 80\% powered study often surprises cognitive psychologists, because in their experience, replicable research can be done with a smaller number. For a long time, samples of 20–24 participants were the norm in experimental psychology \cite{Brysbaert2019}. However, when applying a two-tailed power test with a correlation coefficient of 0.5 \cite{cohen1988spa} and an assumed significance level of $\alpha$=0.05, we found that at least 34 test subjects would be required to achieve 80\% power. Where correlation is notably less, for example stress biomarker correlation with a periodic stress label, substantially more subjects could be required to achieve at least 80\% statistical power. Iqbal \emph{et al.} \cite{Iqbal2022a} specifically performed a power analysis and similarly concluded that at least 34 test subjects would be required to achieve 80\% statistical power, and built their Stress-Predict dataset using 35 test subjects.\\

\noindent Considering the small number of subjects contained in the datasets utilized in the experiments reviewed in this paper (Figure \ref{fig:figure6}), the statistical power of the experiments and subsequent conclusions reached on the accuracy achieved will be overshadowed, more so if these trained models were applied on new, unseen datasets (to confirm generalization). This holds true when the objective is to infer an unknown truth from the observed data, and hypothesis testing provides a specific framework whose inferential target is a binary truth (stressed vs. non-stressed). For example, whether an EDA biomarker from wearable device data provides a signal that correlates with an elevated level of stress. Li \emph{et al.} \cite{Li2020} provides a detailed discussion and guidelines for choosing between the two strategies (hypothesis testing versus machine learning classification) when designing an experiment that can assist researchers in choosing a strategy and when required, validate whether their sample size contains sufficient power.

\subsection{Lack of Generalization}

\begin{figure}[h!]
\centering
\fbox{\includegraphics[width=\textwidth]{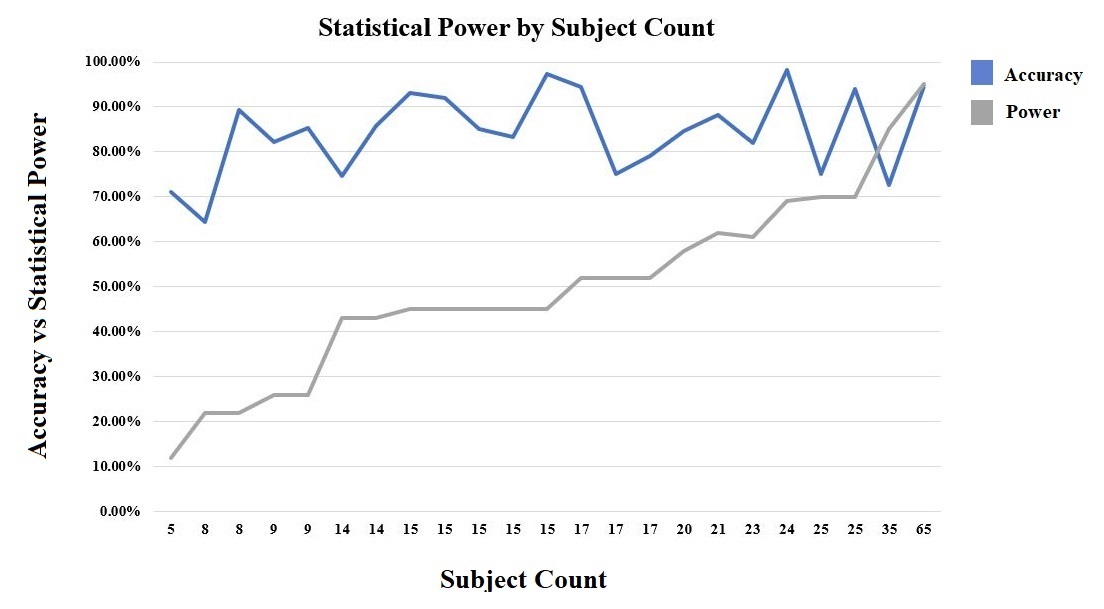}}
\caption{\label{fig:figure6}Accuracy based on number of subjects included in study.}%
\end{figure}

\noindent Interestingly, as shown in Figure \ref{fig:figure6}, there appears to be no obvious correlation between the number of subjects included in the study with the reported accuracy rate. In virtually all the reviewed studies the number of subjects were less than 30. None of these studies apart from Mishra \emph{et al.} \cite{Mishra2020g} and Liapis \emph{et al.} \cite{Liapis2021}, tested generalization of the resulting models on a totally unseen, new dataset, to further validate the reported accuracy achieved in the experiments, when trained on a public dataset.\\

\noindent Figure \ref{fig:figure6} further shows the calculated statistical power given a two-tailed power test with a correlation coefficient of 0.5 \cite{cohen1988spa}, and an assumed significance level of $\alpha$=0.05 for each of the studies reviewed, based on the number of unique test subjects contained within each dataset when used for training and validation. Of these, datasets utilized by Greco \emph{et al.} \cite{Greco2021g} and Ehrhart \emph{et al.} \cite{Ehrhart2022} achieved at least 80\% power by using non-public datasets while the public WESAD \cite{Garg2021, Alshamrani2021g, Iqbal2021g, Indikawati2020g, Schmidt2018, Nkurikiyeyezu2019, Liapis2021} and SWELL \cite{Iqbal2022, Nkurikiyeyezu2019} datasets achieve 45\% and 70\%  power respectively, based on number of test subjects included.\\

\noindent The majority of studies in this review use a custom or public dataset to train their machine learning algorithms using time-series biomarker data within that dataset. These models are then evaluated using the test set of the same dataset, meaning the same experimental setup, and sometimes, different biomarker recordings of previously observed subjects during training. This cannot ensure generalizability of the developed model to other subjects or datasets. Recently some studies are evaluating the potential of person-specific models and their promise in improving generic stress detection models \cite{Nkurikiyeyezu2019}. Of the studies reviews in this paper, and scored using the IJMEDI checklist \cite{Cabitza2021}, three studies \cite{Ehrhart2022, Liapis2021, Greco2021g} were found to likely achieve generalization (Table \ref{tab:ijmedi}), based on model validation and the use of sufficiently large training datasets based on the number of individual study subjects included.\\ 

\FloatBarrier

\begin{figure}[h!]
\centering
\fbox{\includegraphics[width=\textwidth]{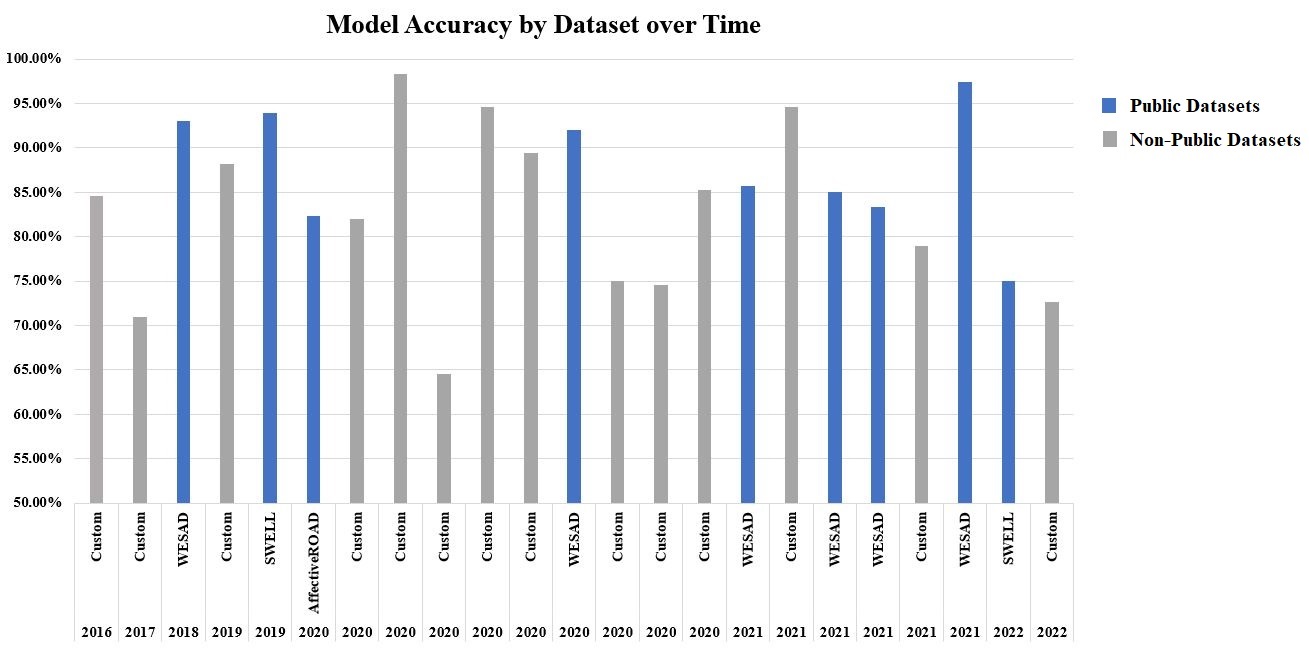}}
\caption{\label{fig:figure7}Reported accuracy based on dataset utilized over time.}%
\end{figure}

\noindent Focusing specifically on results reported when models are trained and evaluated on the most commonly used WESAD dataset (Table \ref{tab:modelsreviewed}), we note that of those experiments reporting accuracy rates higher than 90\% \cite{Indikawati2020g, Schmidt2018, Nkurikiyeyezu2019, Liapis2021}, all included both EDA and HR (or HRV) biomarkers, while those excluding either the HR or EDA biomarkers \cite{Garg2021, Alshamrani2021g, Iqbal2021g} consistently reported accuracy rates below 86\%, irrespective of feature-engineering or cross-validation technique applied. This observation is in line with findings by Schmidt \emph{et al.} \cite{Schmidt2018} who noted their reported highest accuracy rate of 93\% dropped to 88.33\% when excluding the HR biomarker during model training and validation. This may indicate that both EDA and HR (or derivatives including HRV) biomarkers play an equally important role in correlation with perceived elevated levels of stress and requires further examination, considering the small sample of experiments reviewed. Additionally, when considering the advancement of machine learning technologies over the last decade, there appears to be no consistent increase in model performance or reported accuracy over time (Figure \ref{fig:figure7}), indicating that model generalization with respect to stress detection and measurement using machine learning techniques remains a challenge.\\

\FloatBarrier

\subsection{Summary}
\noindent The significant observations from this review are:
\begin{itemize}
  \item Technological improvements in wearable devices have seen a rapid improvement in complexity, ease of use and affordability. This has helped many studies to record and analyze various physiological signals that can be used as biomarkers. 
  \item Sensor biomarkers vary across the wearable devices reviewed, with questions remaining on whether all sensor data can be considered valid and accurate for use in stress detection and measurement, or which biomarkers are the best when measuring stress.
  \item Existing work have predominantly used small datasets acquired in a single experimental setup with varying labeling protocols, bringing into question the statistical power of these small datasets when used for both training and validation.
  \item In the studies reviewed, model validation was performed predominantly using LOSO or K-Fold cross-validation, with no further validation on a completely new, unseen dataset recorded in different experimental conditions using new study participants, leaving the question of model generalization unanswered.
\end{itemize}

\subsection{Challenges and future research directions}

\noindent To achieve reliable machine learning models suitable for real-world monitoring of stress, three formidable challenges should be addressed. 
\begin{itemize}
  \item Varying experimental and labeling protocols influence stress measurement and detection accuracy. To address this challenge, there exists the need for a definitive set of test guidelines when using wearable devices to record biomarker data, including appraisal and scoring methodology. In \cite{Epel2018}, the authors concluded that, the appraisal process critically shapes an individual's response to acute stress, while \cite{RuizRobledillo2013} detected lower EDA biomarker activity in response to episodes of acute stress in caregivers of people with Autism Spectrum Disorder, a potential habituation to stress. These findings support the need for a proper understanding of when wearable devices can and should be used, and potential factors that could affect sensor accuracy.
  \item Measurement accuracy is a major challenge that can significantly affect wearable device data and consequently any stress measurements. One of the main problems with current wearables is significant motion artifacts, which may be reduced by measures for better and more stable placement of the device, or through placing the device on other parts of the body.
  \item Another significant challenge is the lack of large, diverse public datasets built from wearable sensor data that can be utilized to build machine learning models for predicting elevated levels of stress that generalize well to unseen data. 
\end{itemize}

\section{Conclusion}\label{sec:conclusion}
\noindent The main objective in automated stress detection and measurement is to develop a robust, highly accurate machine learning model that can generalizing well on new, unseen data. The review presented here synthesized the literature and presented important information about the previous studies concerned with stress prediction using wearable devices. In particular, we reviewed and analyzed the publicly available stress biomarker datasets used in numerous studies, the machine learning techniques applied, their advantages, limitations and ability to generalize on new, unseen data. We also summarized our point of view on challenges and opportunities in this emerging domain. We believe this review will advance knowledge in the general area of machine learning for stress detection using wearable devices, helping the research efforts move one step closer to realizing effective stress detection and management technology.  

 \bibliographystyle{elsarticle-num} 
 \bibliography{cas-refs}





\end{document}